%% file: main.tex
\documentclass{article}

\usepackage[preprint]{neurips_2026}

\usepackage[utf8]{inputenc}
\usepackage[T1]{fontenc}
\usepackage{hyperref}
\hypersetup{hidelinks}   
\usepackage{url}
\usepackage{booktabs}
\usepackage{amsfonts}
\usepackage{amssymb}
\usepackage{pifont}
\usepackage{graphicx}
\usepackage{float}
\usepackage{wrapfig}
\usepackage{makecell}
\usepackage{listings}      
\usepackage{nicefrac}
\usepackage{microtype}
\usepackage[table]{xcolor}
\usepackage{amsmath}
\usepackage[nameinlink,capitalise,noabbrev]{cleveref}
\usepackage{multirow}
\usepackage{subcaption}
\usepackage{tcolorbox}
\tcbuselibrary{skins}

\usepackage{array}
\usepackage{ragged2e}      
\usepackage{fontawesome5}  
\usepackage{siunitx}       
\definecolor{skyblue}{HTML}{92C5DE}
\definecolor{myblue}{HTML}{6691CD}
\definecolor{hl}{RGB}{220,235,255}   

\newtcolorbox{HighlighterBox}[2][]{
	arc=3.8pt,
	left=5.0pt,
	right=5.0pt,
	bottom=2pt,
	top=2pt,
	colback=skyblue!7.5,
	colframe=skyblue!35,
	boxrule=0.8pt,
	colbacktitle=skyblue!35,
	coltitle=myblue!20!black,
	title=\textbf{#2},
	fonttitle=\bfseries,
	before upper=\justifying,
	#1,
}

\definecolor{ncmetabg}{HTML}{F1F4F7}
\definecolor{ncmetaedge}{HTML}{DCE6F5}
\definecolor{ncmetablue}{HTML}{1877F2}
\definecolor{ncapricot}{HTML}{F7EBDD}

\usepackage{xcolor}
\newif\ifshowcolors
\showcolorsfalse

\ifshowcolors

\else

\fi

\newcommand{\papertitle}{Ground3D-LMM: Fine-Grained 3D Point Grounding and Spatial Reasoning with LMM}
\title{\papertitle}

\author{%
  \bfseries
  Amol Harsh$^{1}$, Zongyan Han$^{1,}$\thanks{Corresponding author: Zongyan.Han@mbzuai.ac.ae}, Jean Lahoud$^{1}$,
  Ye Liu$^{2}$, Rao Muhammad Anwer$^{1}$,\\
  \bfseries
  Hisham Cholakkal$^{1}$, Salman Khan$^{1}$, Fahad Shahbaz Khan$^{1,3}$ \\[4pt]
  \normalfont
  $^{1}$Mohamed bin Zayed University of Artificial Intelligence \quad
  $^{2}$The Hong Kong Polytechnic University \\
  $^{3}$Linköping University
}

\makeatletter
\newcommand{\maketitleboxed}[1]{%
  \par
  \begingroup
    \renewcommand{\thefootnote}{\fnsymbol{footnote}}
    \renewcommand{\@makefnmark}{\textsuperscript{\@thefnmark}}
    \long\def\@makefntext##1{%
      \parindent 1em\noindent
      \hbox to 1.8em{\hss $\m@th ^{\@thefnmark}$}##1%
    }
    \thispagestyle{empty}%
    \begin{tcolorbox}[
      enhanced,
      colback=ncmetabg,
      colframe=ncmetaedge,
      boxrule=0.35pt,
      arc=12pt,
      left=0.55cm, right=0.55cm, top=0.45cm, bottom=0.4cm,
      interior style={shade, shading angle=315,
        left color=white!96!ncmetabg,
        right color=ncmetablue!4!ncapricot!8!ncmetabg},
      before skip=0pt, after skip=0.4em,
      grow to left by=1.5pt, grow to right by=1.5pt,
    ]
      \centering
      {\LARGE\bf \@title\par}%
      \def\And{\end{tabular}\hfil\linebreak[0]\hfil\begin{tabular}[t]{c}\bf\rule{\z@}{24\p@}\ignorespaces}%
      \def\AND{\end{tabular}\hfil\linebreak[4]\hfil\begin{tabular}[t]{c}\bf\rule{\z@}{24\p@}\ignorespaces}%
      \begin{tabular}[t]{c}\bf\rule{\z@}{24\p@}\@author\end{tabular}\par
      \vskip 0.12in
      {\small
        \textcolor{black}{\faGithub}\enspace\textbf{Code:}\enspace
        \href{https://github.com/AmolHarsh/ground3d-lmm}{\textcolor{ncmetablue}{\texttt{github.com/AmolHarsh/ground3d-lmm}}}\\[3pt]
        \textcolor{cyan!60!blue}{\faGlobeAmericas}\enspace\textbf{Project Page:}\enspace
        \href{https://amolharsh.github.io/ground3d-lmm/}{\textcolor{ncmetablue}{\texttt{amolharsh.github.io/ground3d-lmm}}}\\[3pt]
         \textcolor{orange!90!yellow}{\faDatabase}\enspace\textbf{Dataset:}\enspace
        \href{https://huggingface.co/datasets/amolharsh/Ground3D_Dataset}{\textcolor{ncmetablue}{\texttt{huggingface.co/datasets/amolharsh/Ground3D\_Dataset}}}%
      \par}
      \vskip 0.12in
      \captionsetup{justification=justified,singlelinecheck=false,font=small}
      \includegraphics[width=\linewidth]{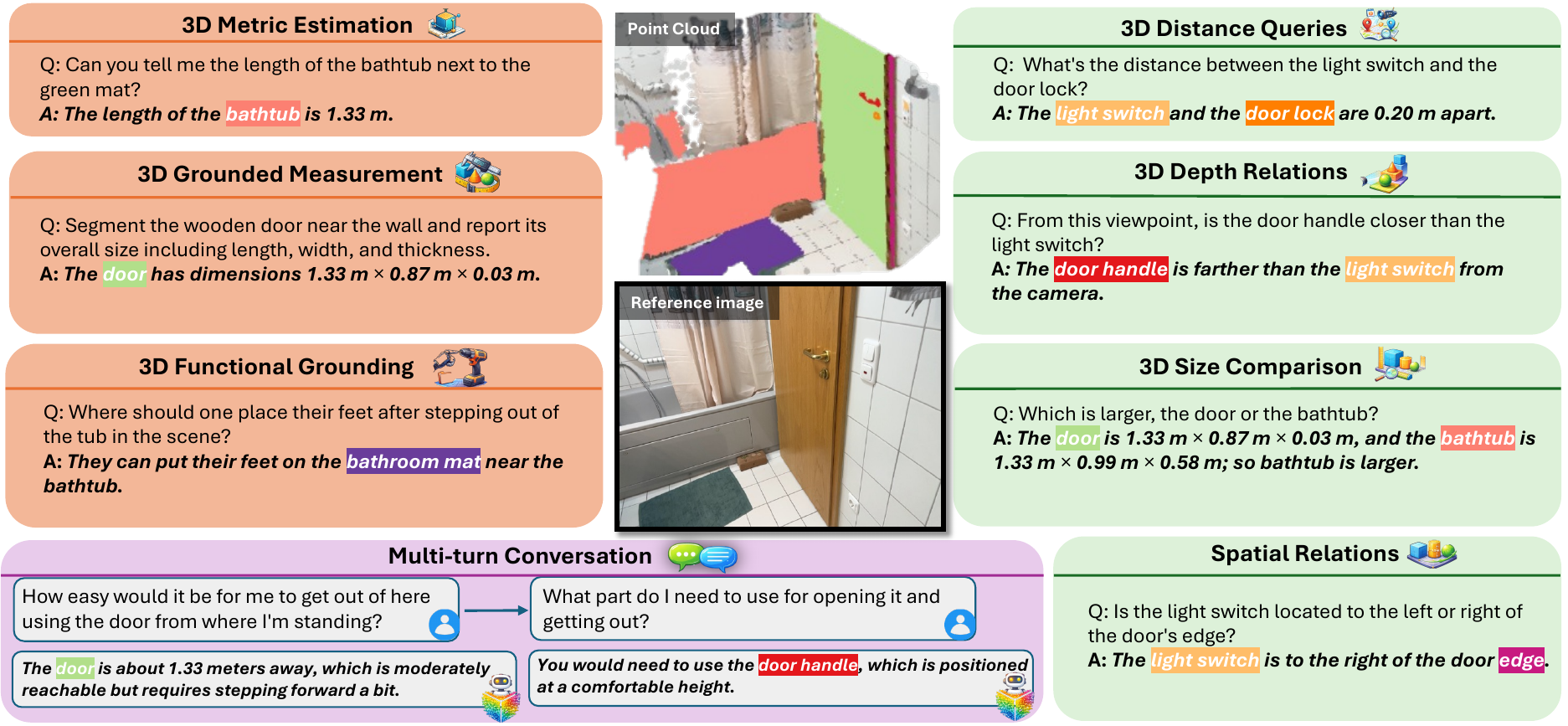}
      \captionof{figure}{Ground3D-LMM answers open-vocabulary object/part questions with grounded (3D segmentation) and metric (real-world units) outputs. We show example queries spanning 3D metric estimation, grounded measurement, functional grounding, spatial relations, and multi-object distance/depth/size reasoning. Queries also include multi-turn dialogue with follow-up questions that reuse previously grounded entities; the model outputs a point-level mask when a response contains a \texttt{<SEG>} trigger.}
      \label{fig:teaser}
      \vskip 0.14in
      \begingroup
        \leftskip=1.5em \rightskip=1.5em
        \centerline{\large\bf Abstract}\vspace{0.6ex}
        \small #1\par
      \endgroup
    \end{tcolorbox}%
    \@thanks
    \@notice
  \endgroup
  \let\maketitle\relax
  \let\thanks\relax
}
\makeatother

\begin{document}

\maketitleboxed{\input{sections/0_abstract}}


\input{sections/1_intro}
\input{sections/2_related}
\input{sections/4_data}
\input{sections/3_method}
\input{sections/5_experiments}

\input{sections/6_conclusion}
\input{sections/7_acknowledgment}

\bibliographystyle{plain}
\bibliography{main}

\clearpage
\appendix

\begin{center}
  {\LARGE\bfseries Supplementary Material of\\[0.2em]
  Ground3D-LMM: Fine-Grained 3D Point\\[0.2em]
  Grounding and Spatial Reasoning with LMM\par}
\end{center}
\vspace{1.5em}

\input{appendix/A_semantic_novelty}
\input{appendix/B_qualitative_results}

\input{appendix/C_additional_quantitative_results}

\input{appendix/C_data_prompts}

\clearpage

\end{document}

%% file: sections/0_abstract.tex
Natural-language queries about 3D environments become actionable when responses are verifiable and metric.
Verifiability requires explicit grounding to the referred 3D region, while metric answers report physical measurements in real-world units (e.g., size, thickness, clearance, and distance).
Existing 3D large multimodal models (LMMs) 
approaches remain limited: conversational systems typically respond without explicit 3D grounding, while 3D grounding models are not designed for interactive, metric-aware dialogue.
In this paper, we present \textbf{Ground3D-LMM}, a unified model that takes a point cloud and an optional RGB image as input and supports 3D spatial conversation with \emph{(i)} point-grounded responses and \emph{(ii)} metric numeric outputs at both object and part granularity, including multi-object queries.
To evaluate this intersection of grounding and measurement, we define the \emph{3D Grounded Measurement} task, which requires predicting the referred 3D region and the corresponding metric quantities in real-world units.
We introduce a large-scale dataset built on ScanNet and ScanNet++ datasets with dense object and part annotations and roughly \( 2.5\)M question-answer pairs spanning eight tasks, along with a manually verified test set.
Extensive experiments on multiple datasets and tasks show that our proposed Ground3D-LMM model provides a strong baseline for grounded, metric-aware 3D conversational understanding.
Our dataset and model are publicly available.

%% file: sections/1_intro.tex
\section{Introduction}
\label{sec:intro}

Natural interaction with a 3D environment often requires two capabilities.
First, the system must verify what it is talking about by grounding language to the correct region in the scene.
Second, it must be able to provide metric measurements: answers such as "the chair is larger" are less useful than answers expressed in meters or centimeters.
These requirements appear in various tasks, such as checking clearances in robotics, measuring surfaces in AR/VR, and locating functional parts of objects in assistive agents.

Current 3D vision-language models rarely satisfy both requirements in one system.
Some works focus on 3D conversation or question answering, but provide text without explicit point-level grounding.
Other works produce 3D masks or boxes from text, but do not support interactive dialogue and are not designed to return metric quantities.
Moreover, open-vocabulary part-level reasoning is still underexplored, even though many real requests refer to functional parts (e.g., seat of a chair, handle of a drawer).

In this paper, we propose Ground3D-LMM, a unified model that supports grounded and metric-aware 3D conversation.
Given a colored point cloud, a user query, and an optional RGB image, our model predicts a textual response along with a 3D segmentation mask for the referred object or part.
Moreover, measurements are returned in metric physical units and support multi-object queries (e.g., comparing two chairs, or measuring the clearance between a desk and a drawer).
Figure \ref{fig:teaser} shows examples of the open-vocabulary object/part queries with grounded and metric outputs.
Our key idea is to embed 3D point features as tokens alongside text (and optional image tokens) in an LMM, so the model can reason jointly over language and 3D geometry. When the model generates a special grounding trigger (e.g., \texttt{<SEG>}) tied to a phrase, a lightweight segmentation head predicts the corresponding point-level mask.

The key contributions of our work are as follows:
\begin{itemize}
    \item We present \textbf{Ground3D-LMM}, a unified point-cloud LMM that produces text and 3D masks in a single conversational interface, and supports metric measurements at object and part granularity.
    \item We introduce a large-scale dataset with dense 3D object/part annotations and multi-turn conversations designed for grounded and metric reasoning, along with a curated evaluation split with manual verification.
    \item We formalize \emph{3D Grounded Measurement}, a task that requires predicting a referred 3D region and metric quantities under a single evaluation protocol.
    \item We conduct extensive experiments on our proposed dataset and relevant subtask benchmarks, demonstrating that Ground3D-LMM consistently achieves superior performance across all tasks.
\end{itemize}

%% file: sections/2_related.tex
\section{Related Work}\label{sec:related_work}



\newcommand{\cmark}{{\color{green!60!black}\ding{51}}} 
\newcommand{\xmark}{{\color{red!75!black}\ding{55}}}   
\definecolor{oursbg}{RGB}{237,232,245}

\begin{table}[t]
\caption{
\textbf{Comparison of recent 3D multimodal / conversational models.} We compare input modalities and whether a method supports \emph{3D spatial conversation with metric values}, \emph{point-grounded conversational outputs} (text paired with point-level masks), and grounding granularity. "seg" denotes point-level masks and "num" denotes metric numeric outputs in real units.
}
\centering
\normalsize              
\setlength{\tabcolsep}{4pt}
\renewcommand{\arraystretch}{1.18}

\resizebox{0.98\linewidth}{!}{%
\begin{tabular}{ll l c c c c c c c c}
\toprule
\multirow{2}{*}{Year} &
\multirow{2}{*}{Venue} &
\multirow{2}{*}{Model / Paper} &
\multicolumn{3}{c}{Input} &
\multirow{3}{*}{\shortstack{3D \\ Spatial \\ Conv.}} &
\multirow{3}{*}{\shortstack{Point Grounded \\ Conv.\\(Text + Seg)}} &
\multicolumn{3}{c}{Grounding Granularity} \\
\cmidrule(lr){4-6}\cmidrule(lr){9-11}
& & &
RGB &
\shortstack{Point \\ Cloud}&
\shortstack{Scene \\ level} &
& &
\shortstack{Object\\(seg / num)} &
\shortstack{Part\\(seg / num)} &
\shortstack{Multi \\ Object} \\
\midrule

2025 & CVPR    & Inst3D-LLM~\cite{yu2025inst3d}              & \cmark & \cmark & \cmark & \xmark & \xmark & \xmark & \xmark & \cmark \\
\rowcolor{gray!15}
2025 & NeurIPS & SD-VLM~\cite{chen2025sd}                  & \cmark & \xmark & \cmark & \cmark & \xmark & \cmark & \xmark & \cmark \\
2025 & NeurIPS & MLLM-For3D~\cite{huang2025mllmfor3d}              & \cmark & \xmark & \cmark & \xmark & \xmark & \cmark & \xmark & \xmark \\
\rowcolor{gray!15}
2025 & ICCV    & Kestrel~\cite{ahmed2025kestrel}                 & \xmark & \cmark & \xmark & \xmark & \cmark & \xmark & \cmark & \xmark \\
2025 & CVPR    & SeeGround~\cite{li2025seeground}               & \xmark & \cmark & \cmark & \xmark & \xmark & \xmark & \xmark & \xmark \\
\rowcolor{gray!15}
2024 & ICLR    & Reason3D~\cite{huang2024reason3d}                & \xmark & \cmark & \cmark & \xmark & \xmark & \cmark & \xmark & \xmark \\
2024 & ECCV    & SegPoint~\cite{he2024segpoint}                & \xmark & \cmark & \cmark & \xmark & \xmark & \cmark & \xmark & \cmark \\
\rowcolor{gray!15}
2024 & ICLR    & Open-YOLO 3D~\cite{boudjoghra2025open}            & \cmark & \cmark & \cmark & \xmark & \xmark & \cmark & \xmark & \xmark \\
2024 & CVPR    & Open3DIS~\cite{nguyen2024open3dis}                & \xmark & \cmark & \cmark & \xmark & \xmark & \cmark & \xmark & \xmark \\
\rowcolor{gray!15}
2024 & ECCV    & PARIS-3D~\cite{kareem2024paris3d}               & \xmark & \cmark & \xmark & \xmark & \cmark & \xmark & \cmark & \xmark \\
2023 & CVPR    & PartSLIP~\cite{liu2023partslip}                & \xmark & \cmark & \xmark & \xmark & \xmark & \xmark & \cmark & \xmark \\
\midrule

\rowcolor{oursbg}
\multicolumn{3}{l}{\textbf{Ours}} &
\cmark & \cmark & \cmark & \cmark & \cmark & \cmark & \cmark & \cmark \\
\bottomrule
\end{tabular}%
}

\label{tab:3d_mllm_comparison}
\end{table}

We review prior work on 3D language grounding, open-vocabulary 3D segmentation, 3D multimodal LLMs, and metric spatial reasoning.
Table~\ref{tab:3d_mllm_comparison} summarizes representative recent systems and highlights which components are typically missing for grounded, metric-aware 3D conversation (e.g., point-level grounding, part granularity, and multi-object measurement).

\paragraph{Language grounding in 3D.}
3D referring expression comprehension and grounding benchmarks require localizing a target object in a scene, usually with a box or a mask.
Benchmarks such as ScanRefer~\cite{chen2020scanrefer} and ReferIt3D~\cite{achlioptas2020referit3d} established the importance of explicit grounding for verifiable 3D interaction.
Recent systems in this direction include Kestrel~\cite{ahmed2025kestrel}, PARIS-3D~\cite{kareem2024paris3d}, SegPoint~\cite{he2024segpoint}, and SeeGround~\cite{li2025seeground}.
Referring segmentation methods such as InstanceRefer~\cite{Yuan_2021_ICCV}, BUTD-DETR~\cite{butd_detr}, EDA~\cite{wu2023eda}, and 3D-STMN~\cite{3dstmn} further improve localization quality, while Multi3DRefer~\cite{zhang2023multi3drefer} studies multi-object referring expressions.
These methods improve localization and grounding quality, but they typically do not support interactive, multi-turn dialogue or metric outputs.

\paragraph{Open-vocabulary 3D segmentation.}
Recent open-vocabulary methods extend 2D vision-language pre-training to 3D point clouds and enable class-agnostic segmentation.
Representative approaches include Open3DIS~\cite{nguyen2024open3dis}, Open-YOLO~3D~\cite{boudjoghra2025open}, PartSLIP~\cite{liu2023partslip}, as well as UniSeg3D~\cite{xu2024unified}.
While these models are strong at producing masks, they are not designed to hold a conversation or provide metric answers with unit consistency, especially at the part level.

\paragraph{3D question answering and multimodal LLMs.}
3D QA datasets and 3D LMMs enable natural-language interaction with scenes.
Popular benchmarks include ScanQA~\cite{azuma2022scanqa} and SQA3D~\cite{ma2022sqa3d}.
Recent LMM-style systems include Inst3D-LLM~\cite{yu2025inst3d}, $S^{2}$-MLLM\cite{xu2025s} ,  MLLM-For3D~\cite{huang2025mllmfor3d}, SD-VLM~\cite{chen2025sd}, and Reason3D~\cite{huang2024reason3d}.
However, most systems answer in text only, which makes it hard to verify whether the model is referring to the correct region.
In addition, metric reasoning is often treated as a side task and is not standardized across benchmarks.

\paragraph{Metric spatial reasoning.}
Several works study size, distance, and spatial relations in either images or 3D data.
N3D-VLM~\cite{wang2025n3d} and SD-VLM~\cite{chen2025sd} move toward metric reasoning in 3D settings, but metric evaluation is not consistently paired with explicit point-level grounding of the referred region.
SpatialVLM \cite{chen2024spatialvlm} facilitates basic spatial query responses from single images, and SpatialRGPT \cite{cheng2024spatialrgpt} advances this by incorporating 3D scene-graph-derived regional awareness. 
Nevertheless, both frameworks remain constrained by a reliance on 2D-centric training data or external depth plugins, ultimately failing to achieve a native, metric-scale 3D understanding of physical distances and object sizes.
These efforts highlight the need for numeric supervision, but they do not provide a unified setting that requires grounding a referred region and returning metric quantities for that region.

On the other hand, our proposed Ground3D-LMM method combines point-level grounding and metric reasoning in a single conversational interface and emphasizes open-vocabulary part understanding.
We also provide a task and dataset that standardize evaluation for grounding and measurement.

%% file: sections/4_data.tex
\section{Ground3D Dataset}\label{sec:data}

\begin{figure}[t]
    \centering
    \includegraphics[width=1\linewidth]{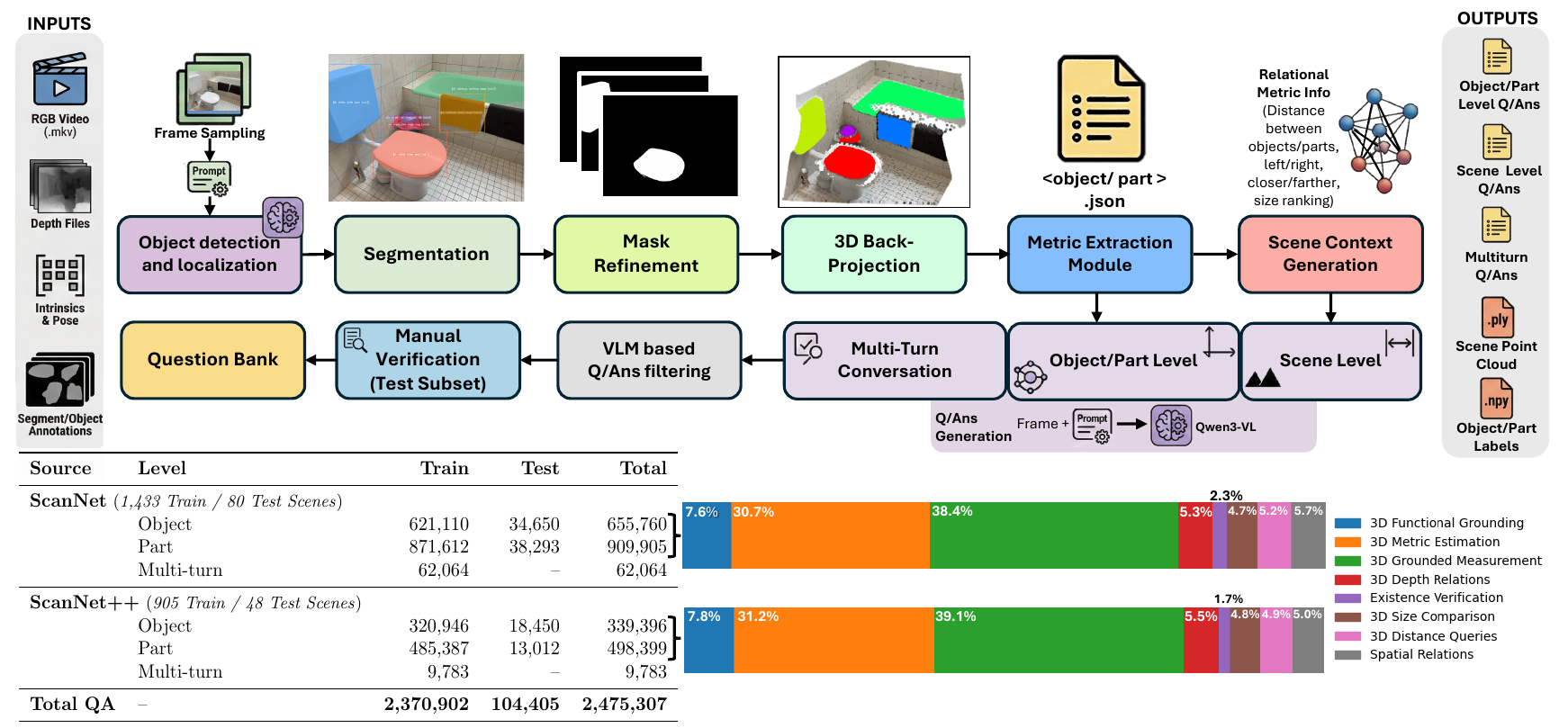}
    \caption{\textbf{Ground3D statistics and annotation pipeline.}
    Starting from ScanNet~\cite{dai2017scannet} and ScanNet++~\cite{yeshwanth2023scannet++} RGB-D sequences, we sample representative frames, detect object/part candidates, generate and refine 2D masks, lift them to point-level 3D supervision via depth back-projection, and compute metric attributes and spatial relations. We then synthesize grounded QA at object/part level, scene-relational level, and multi-turn dialogue, followed by automatic filtering and manual verification on an evaluation subset.}
    \label{fig:annotation}
\end{figure}

\paragraph{Task Definition.} 
We define \emph{3D Grounded Measurement} as follows: given a 3D scene (point cloud, optionally with an aligned RGB view) and a natural-language query, the model must
(i) identify the referred object or part as a point-level 3D mask, and
(ii) answer with the requested physical quantity in consistent units (e.g., dimensions, thickness, clearance, or inter-object distance).
Dataset statistics and the annotation pipeline are summarized in \Cref{fig:annotation}.

\subsection{Dataset Subtasks}
Ground3D dataset is organized as a set of complementary subtasks that cover grounded conversation and metric reasoning at both object and part granularity:
\textbf{3D metric estimation:} report object/part dimensions derived from the referred region;
\textbf{3D grounded measurement:} jointly return a mask and the requested measurement for the same referent;
\textbf{3D distance queries:} compute distances/clearances between two grounded regions;
\textbf{3D depth relations:} reason about closer/farther and front/behind based on 3D geometry;
\textbf{3D size comparison:} compare sizes across objects/parts using consistent units; and
\textbf{Spatial relations:} predict left/right/above/below relations grounded to explicit regions.

We also include subtasks that reflect interactive usage, where the model must handle absence, affordances, and context carried across turns:
\textbf{Existence verification:} confirm the presence or absence of a queried object, including plausible but non-existent queries to reduce false positives;
\textbf{Functional grounding:} identify parts that support an action or affordance (e.g., "the graspable handle") and ground them; and
\textbf{Multi-turn conversation:} maintain context across turns while alternating between grounded and purely textual responses.






\subsection{Dataset Construction}
We use around 2.5K scenes from ScanNet and ScanNet++ datasets to build Ground3D dataset and benchmark.
Each scene provides calibrated RGB-D frames with camera intrinsics and poses, enabling us to lift 2D segmentations into 3D and compute metric quantities in real units. An overview of the pipeline is shown in \Cref{fig:annotation}.

\paragraph{Stage 1: Frame sampling and 2D object/part grounding.}
For each scene, we select a fixed number of sharp, well-spread frames to cover diverse viewpoints. We prompt Qwen3-VL-30B with a structured instruction to propose visible objects and semantically meaningful parts, returning object/part names and bounding boxes in JSON format. The prompt follows a hierarchical object$\rightarrow$part strategy and allows open-vocabulary candidates to increase part diversity beyond the native dataset labels.

\paragraph{Stage 2: 2D mask generation and refinement.}
Given the predicted boxes, we generate instance masks with SAM2~\cite{ravi2024sam}. We refine masks in two ways. When a prediction can be aligned to ScanNet/ScanNet++ instances, we project masked pixels into 3D using depth and pose and keep pixels consistent with the dominant 3D instance (with name-to-label matching for the predicted category). For candidates that do not correspond to a known label, we apply a geometry-only refinement that erodes uncertain boundaries (distance-transform based), retaining a stable core region and reducing boundary noise.

\paragraph{Stage 3: 3D back-projection and mask cleanup.}
We back-project depth into a camera-frame point cloud and lift each refined 2D mask to an object/part-specific 3D point subset. To suppress depth noise and accidental leakage, we remove 3D outliers using per-axis percentiles (e.g., 5\%-95\%), producing clean point-level masks used for supervision and evaluation.

\paragraph{Stage 4: Metric and relational context generation.}
From each grounded point set, we compute per-instance geometric attributes including camera-frame centroid, oriented bounding box (OBB) dimensions (length, width, thickness), and camera distance. We additionally compute scene-level relations used by relational subtasks: pairwise distances/clearances (5th-percentile closest-point distance), left/right ordering (camera-$x$), closer/farther (camera-$z$), and global size ranking via OBB volume. All metrics are defined over the visible 3D geometry from the current viewport rather than the full object mesh, matching what the model perceives and reflecting embodied and robotics settings.

\paragraph{Stage 5: Grounded Q/A synthesis (three levels).}
We synthesize question--answer pairs with Qwen3-VL-8B conditioned on the computed attributes. \textbf{Level 1} generates single-object/part Q/A covering metric estimation, grounded measurement, and functional grounding. \textbf{Level 2} samples ordered object/part pairs to generate distance, depth, size, spatial, and existence queries conditioned on pairwise relations. \textbf{Level 3} assembles three-turn grounded dialogues for selected object--part pairs, with motives such as reachability, comparison, identification, organization, and preference. Across all levels, grounded spans are marked with \texttt{<p>...\ </p><SEG>} to align answer text with the corresponding point mask(s).


\paragraph{Verification and filtering.}
We apply rule-based cleanup to remove malformed outputs (e.g., broken \texttt{<p>...\ </p><SEG>} tags, missing numbers/units, or inconsistent metrics), prune noisy segmentations, and discard corrupted depth/metric entries. A secondary VLM-based verifier (Qwen3-VL-30B) checks (a) phrase--mask consistency, (b) metric plausibility against the image evidence, and (c) answer contradictions, dropping any item that fails. Finally, we manually inspect all 128 evaluation scenes for mask quality and metric correctness (${\sim}26\%$ removed); a further ${\sim}3\%$ is removed by VLM filtering, yielding 104{,}405 retained high-quality samples. The resulting dataset includes object-level, part-level, scene-level, and multi-turn grounded Q/A pairs together with aligned 3D masks and metric annotations.






%% file: sections/3_method.tex
\section{Method}\label{sec:method}

\begin{figure}[t]
    \centering
    \includegraphics[width=1\linewidth]{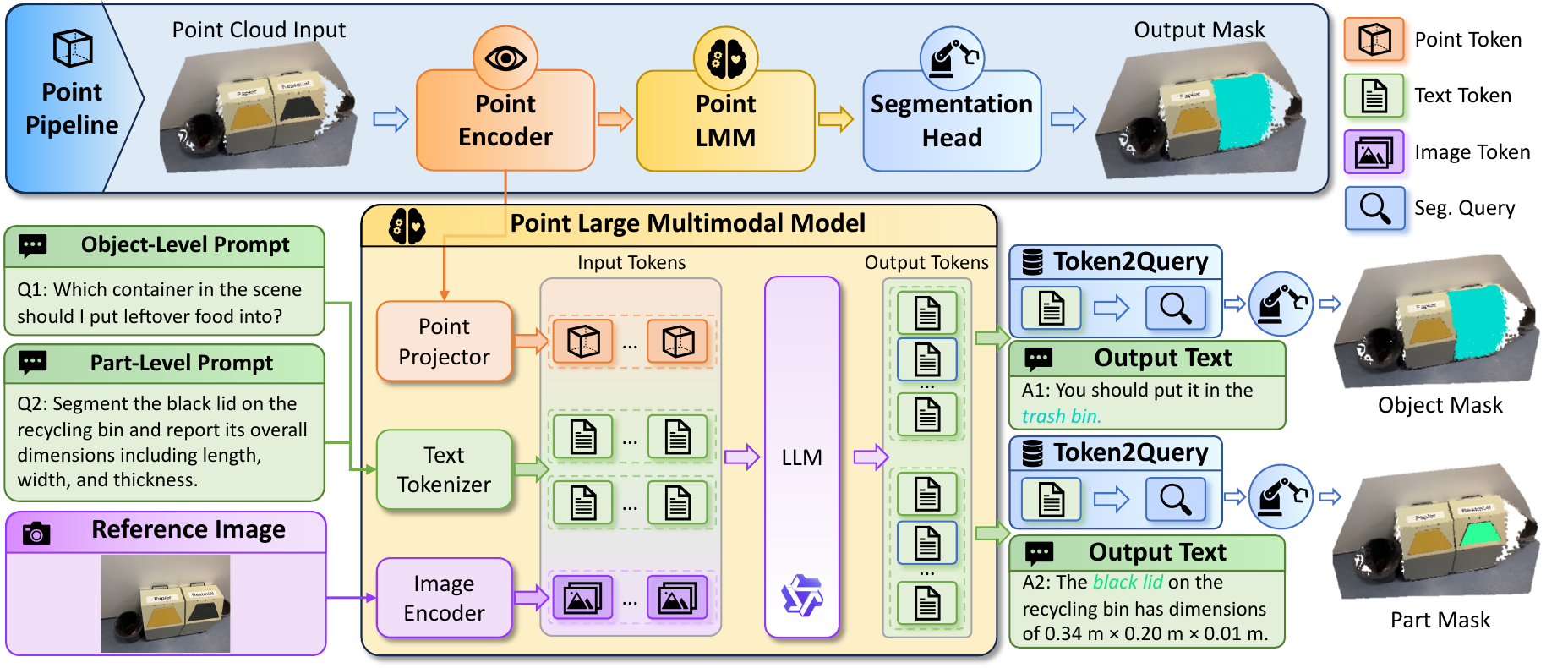}
    \caption{\textbf{Ground3D-LMM overview.} Given a colored point cloud and an optional RGB image, the model answers in text and outputs point-level masks when \texttt{<SEG>} is triggered.}
    \label{fig:method}
\end{figure}

\paragraph{Problem Definition.}
We represent a colored point cloud of \( L \) points as \(X \in \mathbb{R}^{L \times 6}\), where each point has coordinates and color \([x,y,z,r,g,b]\).
Given a natural-language query \(q\) and an optional RGB image \(I\) our goal is to produce:
(i) a textual answer \(y\), and (ii) when the query refers to a specific region, a point-level mask \(m \in \{0,1\}^{N}\) that segments the referred object or part.
For metric queries, the answer also includes numeric values (e.g., length, width, height, thickness, distances) in physical units.


\subsection{Ground3D-LMM Overview}
The overall framework of our Ground3D-LMM is as shown in~\Cref{fig:method}. 
Ground3D-LMM couples a 3D point encoder with a large multimodal language model.
The point encoder produces a set of point or superpoint features.
The LMM performs instruction following and generates the response text.
When the output contains a special grounding trigger token (\texttt{<SEG>}), we run a segmentation head that predicts a 3D mask aligned with the triggered phrase.
Moreover, the LLM computes metric quantities for the predicted region (e.g., oriented bounding box dimensions, distances between regions), and these values are verbalized in the final answer.

\subsection{Point Encoder}
To balance computational efficiency with the retention of fine-grained spatial details, we employ a voxelization-based preprocessing strategy~\cite{choy20194d}. Specifically, the input point cloud consisting of $L$ points is downsampled into $M$ superpoints ($M < L$), which serves to aggregate local geometric structures while filtering out redundant noise. These superpoints are subsequently processed by a sparse 3D U-Net to extract hierarchical point-wise features. The resulting representation, $\mathcal{F}_p \in \mathbb{R}^{M \times d}$ (where $d$ denotes the embedding dimensionality), captures both local geometry and global scene context, providing a robust foundation for subsequent multimodal alignment.

\subsection{Point Large Multimodal Model (Point LMM)}
The core of our framework is a Large Multimodal Model (LMM) capable of reasoning across 3D, 2D, and textual modalities. 

\paragraph{Multimodal Alignment.} To bridge the modality gap, we utilize a linear projector $P$ to map the 3D point features into the LLM's joint latent space:
\begin{equation}
	\mathcal{T}_p = P(\mathcal{F}_p).
\end{equation}
These point tokens $\mathcal{T}_p$ encapsulate essential spatial priors, enabling the LLM to perceive the 3D environment. To further augment the model's grounding capability, we incorporate a camera-view reference image. The extracted image tokens $\mathcal{T}_{img}$ provide complementary visual cues (e.g., relative spatial position in the instruction) that are often ambiguous in raw point clouds, thereby facilitating more precise spatial reasoning.

\paragraph{Instruction Formulation.} 
We follow a prompt-based learning paradigm where the model is tasked with answering scene-related queries regarding objects or their constituent parts. Specifically, we reserve $M$ special tokens, denoted as \texttt{<|point|>}, within the input instruction. These placeholders are dynamically replaced by the embedded point tokens $\mathcal{T}_p$. To enable task-specific output, we adopt the segmentation-aware signaling mechanism where target objects or parts are wrapped within specific tags (e.g., \texttt{<p>object\_name</p><SEG>}). 

The final response generation is formulated as:
\begin{equation}
	\mathcal{Y}_{out} = \text{LMM}(\mathcal{T}_{text}, \mathcal{T}_{img}, \mathcal{T}_p),
\end{equation}
where $\mathcal{T}_{text}$ represents the tokenized user instruction as below.

\begin{HighlighterBox}{\footnotesize Prompt Example}
	\ttfamily\scriptsize 
	\noindent You are an AI visual assistant analyzing a 3D point-cloud scene. Your task is to provide spatial analysis based on the point-cloud data and the given image.
	Please include in your response the objects or object parts annotated with <p>object name</p><SEG> or <p>part name</p><SEG>.
	
	\noindent The point-cloud data is  \textless|point|\textgreater{}\dots\textless|point|\textgreater{}.
	
	\noindent The question is: Which container in the scene should I put leftover food into?
\end{HighlighterBox}

\subsection{Segmentation Head}
Upon generating the textual response, we isolate the latent representations associated within the segmentation tokens, which is denoted as $\mathbf{t}$. We employ a query embedding function $Q$ to transform these states into learnable segmentation queries $\mathbf{q}$. These queries interact with the dense point features $\mathcal{F}_p$ through a sequence of attention-based refinements:


\begin{equation}
    \begin{aligned}
        \mathbf{q}^{(0)} &= Q(\mathbf{t}), & \qquad \mathbf{q}^{(1)} &= \text{Cross-Attn}(\mathbf{q}^{(0)}, \mathcal{F}_p), \\
        \mathbf{q}^{(2)} &= \text{Self-Attn}(\mathbf{q}^{(1)}), & \qquad \mathbf{q}^{(f)} &= \text{FFN}(\mathbf{q}^{(2)}).
    \end{aligned}
\end{equation}

Finally, a lightweight mask decoder computes the similarity between the refined query $q^{(f)}$ and the point features $\mathcal{F}_p$ to produce the final binary mask:
\begin{equation}
	\mathcal{M} = \text{Decoder}(q^{(f)}, \mathcal{F}_p).
\end{equation}

\subsection{Training Objectives}
The Ground3D-LMM is trained end-to-end using a multi-task objective function that encompasses both spatial grounding and linguistic generation. The total loss $\mathcal{L}_{total}$ is defined as a weighted sum of the segmentation loss $\mathcal{L}_{seg}$ and the language modeling loss $\mathcal{L}_{text}$:
\begin{equation}
	\mathcal{L}_{total} = \lambda_{seg} \mathcal{L}_{seg} + \lambda_{text} \mathcal{L}_{text}
\end{equation}
where $\lambda_{seg}$ and $\lambda_{text}$ are hyperparameters balancing the two objectives.

\paragraph{Segmentation Loss.} To ensure precise mask prediction for both objects and parts, we utilize a combination of Binary Cross-Entropy (BCE) loss and Dice loss~\cite{milletari2016v} for the segmentation head. While the BCE loss enforces  point-wise classification accuracy, the Dice loss is employed to mitigate the class imbalance issue inherent in sparse point clouds:
\begin{equation}
	\mathcal{L}_{seg} = \mathcal{L}_{BCE}(\mathcal{M}, \mathcal{M}_{gt}) + \mathcal{L}_{Dice}(\mathcal{M}, \mathcal{M}_{gt})
\end{equation}
where $\mathcal{M}$ and $\mathcal{M}_{gt}$ denote the predicted mask and the corresponding ground-truth label, respectively.

\paragraph{Language Modeling Loss.} For the textual response generation, we adopt the standard auto-regressive cross-entropy loss. The model is optimized to predict the next token in the sequence given the multimodal context:
\begin{equation}
	\mathcal{L}_{text} = -\sum_{i=1}^N \log P(y_i \mid y_{<i}, \mathcal{T}_{text}, \mathcal{T}_{img}, \mathcal{T}_p)
\end{equation}
where $y_i$ represents the $i$-th token in the target output sequence $\mathcal{Y}_{gt}$ of length $N$. 
This loss ensures that the LLM maintains its reasoning and conversational capabilities while learning to trigger the segmentation task via special tokens.

%% file: sections/5_experiments.tex
\section{Experiments}\label{sec:experiments}


\subsection{Evaluation Metrics}

\paragraph{Segmentation}
We evaluate point grounding using mean IoU (mIoU) between predicted and ground-truth point masks.

\paragraph{Metric answers}

For queries requiring numerical responses, we extract predicted scalar values using the LLM and normalize units (e.g., cm, m) to a common scale.  
We report Absolute Percentage Error (APE), defined as 
$\mathrm{APE} = \left|\frac{\hat{s}-s}{s}\right| \times 100\%$, 
as well as the $\delta$ success rate, where 
$\delta = \max\!\left(\frac{\hat{s}}{s}, \frac{s}{\hat{s}}\right)$ and a prediction is considered correct if $\delta \le \tau$ (with $\tau=1.25$). Here, $s$ denotes the ground-truth physical quantity and $\hat{s}$ the predicted value.

\paragraph{Grounded Measurement.}
To jointly assess grounding and metric accuracy, we introduce the \emph{Grounded-Measurement Success Rate} (GM-$\delta$). A prediction is counted as successful only when its mask IoU with the ground truth exceeds $0.3$ \emph{and} its metric prediction satisfies $\delta \le 1.25$. This unified metric integrates segmentation quality and metric correctness into a single score; results are reported in the supplementary material.

\paragraph{Text quality}
To compare text outputs across models, we report a rubric-based evaluation over two criteria:
\emph{Hallucination} (penalizing unsupported claims about the scene), and \emph{Completeness} (whether the response addresses all requested quantities). Each criterion is scored on a 0--10 scale. Scores are averaged over the Ground3D evaluation split; the same judging protocol is applied to all methods.


\subsection{Experimental Setup}
We train Ground3D-LMM on our dataset and evaluate it on both object-level and part-level tasks. We further evaluate the model on the Reason3D dataset and ScanRefer with additional fine-tuning. 
For the Ground3D dataset, we adopt our predefined data splits for all experiments.
For other datasets, we follow the corresponding splits for training and finetuning. 
We report results for both 3D-only (point cloud) and 3D+2D (point cloud + a reference RGB view) settings.

\paragraph{Implementation Details.} 
For point encoder, we adopt a sparse convolutional backbone~\cite{spconv2022} integrated with superpoint pooling~\cite{robert2023spt}, which is initialized with pre-trained weights from SSTNet~\cite{liang2021instance}. 
For the LMM, we leverage the Qwen3-VL-4B-Instruct~\cite{bai2025qwen3}. 
Both the Point Projector and the query embedding function are implemented as Multi-Layer Perceptrons (MLPs).
Our framework is optimized using the AdamW~\cite{loshchilov2019decoupled} optimizer with a weight decay of 0.05. 
To accommodate the varying convergence rates of different components, we apply a stratified learning rate strategy: the point encoder is fine-tuned with an initial rate of $1 \times 10^{-6}$, the LMM backbone is updated at $1 \times 10^{-5}$, while all other task-specific modules (e.g., projectors and segmentation head) are initialized at $1 \times 10^{-4}$. 
We employ a polynomial learning rate scheduler with a power of 0.95 and set the batch size to 1. 
We utilize LoRA~\cite{hu2022lora} for the language model part. 
The loss weights are unified to $\lambda = 1.0$ without the need for extensive tuning.


\subsection{Quantitative Results}

\input{tables/table_ourdata_split}

\paragraph{Baseline Methods}
We compare our Ground3D-LMM with 3 main baseline: Image baseline, UniSeg3D~\cite{xu2024unified} and Reason3D~\cite{huang2024reason3d}. 
We construct an image-only baseline by providing the corresponding RGB frame for each question to Qwen-VL, which generates textual and metric answers with grounded segmentation phrases. The predicted phrases are localized using Grounded Sam~\cite{ren2024groundedsamassemblingopenworld} to obtain 2D masks, which are then back-projected into 3D by mapping mask pixels through the depth image using camera intrinsics.
UniSeg3D~\cite{xu2024unified} supports open-vocabulary 3D segmentation using text prompts as input.
In contrast, Reason3D~\cite{huang2024reason3d} performs reasoning-based segmentation, where the prompts do not explicitly specify category names but instead require the model to infer the target class.

\paragraph{Results on Object Segmentation}
We report object-level segmentation results on our Ground3D dataset in Table~\ref{tab:object_segmentation}.  
Our Ground3D-LMM consistently demonstrates stronger performance than all baselines. Furthermore, incorporating image information brings additional performance gains, enabling Ground3D-LMM to achieve the best results across all tasks.
We observe that Reason3D~\cite{huang2024reason3d}, which is designed for semantic segmentation, consistently underperforms on our dataset. 
This is because our tasks require precise identification of a specific individual object. Moreover, some tasks involve multiple objects from different classes, a setting for which Reason3D~\cite{huang2024reason3d} is not well suited.

\paragraph{Results on Part Segmentation}
We further evaluate the methods on more fine-grained part segmentation tasks, with the results reported in Table~\ref{tab:part_segmentation}. 
As expected, the performance of all methods drops compared to object-level segmentation, since part segmentation is inherently more challenging and requires finer-grained 3D perception. 
Nevertheless, our method still achieves the best performance on this task. Even when using only point cloud input, our approach outperforms all baselines across all tasks. 

\paragraph{Results on Text Response}
In addition to mask quality, we evaluate response quality and metric accuracy in Table \ref{tab:text_metric_eval}. Across Ground3D-ScanNet and Ground3D-ScanNet++ evaluation subsets, Ground3D-LMM consistently outperforms the image-only baseline and SD-VLM~\cite{chen2025sd}, with further gains when combining 3D and 2D cues. Similar improvements are obeserved on both subsets, demonstrating strong cross-dataset generalisation. In text and metric evaluation, our model achieves lower Mean APE and higher $\delta$ success rate, indicating more accurate quantitative reasoning and improved geometric grounding compared to purely image-based approaches.

\paragraph{Results on Reason3D}
Table \ref{tab:reason3d} compares Ground3D-LMM to Reason3D~\cite{huang2024reason3d} and MLLM-For3D~\cite{huang2025mllmfor3d} on ScanNet scenes following the Reason3D~\cite{huang2024reason3d}  protocol.
Using only point cloud input, our method achieves the best performance, 
surpassing Reason3D~\cite{huang2024reason3d} by a large margin (5.15\% on mIoU). 
When further incorporating image information, Ground3D-LMM consistently outperforms all prior methods and delivers the strongest overall results. 
Notably, even without image inputs and using only a 4B backbone, Ground3D-LMM still significantly outperforms MLLM-For3D equipped with a 7B model and image inputs, demonstrating the effectiveness and efficiency of our design.


\input{tables/table_reason3d}
\paragraph{Results on ScanRefer}
We further evaluate open-vocabulary grounding on ScanRefer~\cite{chen2020scanrefer}, which poses a more challenging setting than Reason3D~\cite{huang2024reason3d}. 
Unlike category-level semantic segmentation, ScanRefer demands instance-level grounding, where the model must localize and segment the unique target object described by a referring expression. 
This requires not only semantic understanding but also fine-grained discrimination among multiple objects of the same category.

With only 3D input, our method already surpasses existing approaches by nearly 20\%, demonstrating strong instance-level reasoning capability from pure point cloud features. When image inputs are incorporated, performance is further improved, exceeding the second-best method by 10\%. 
These results indicate that our approach is effective not only for question-driven semantic segmentation, but also for more challenging single-instance open-vocabulary grounding. 

\input{tables/tabel_scanrefer}
\begin{figure}[t]
    \centering
    \includegraphics[width=1\linewidth]{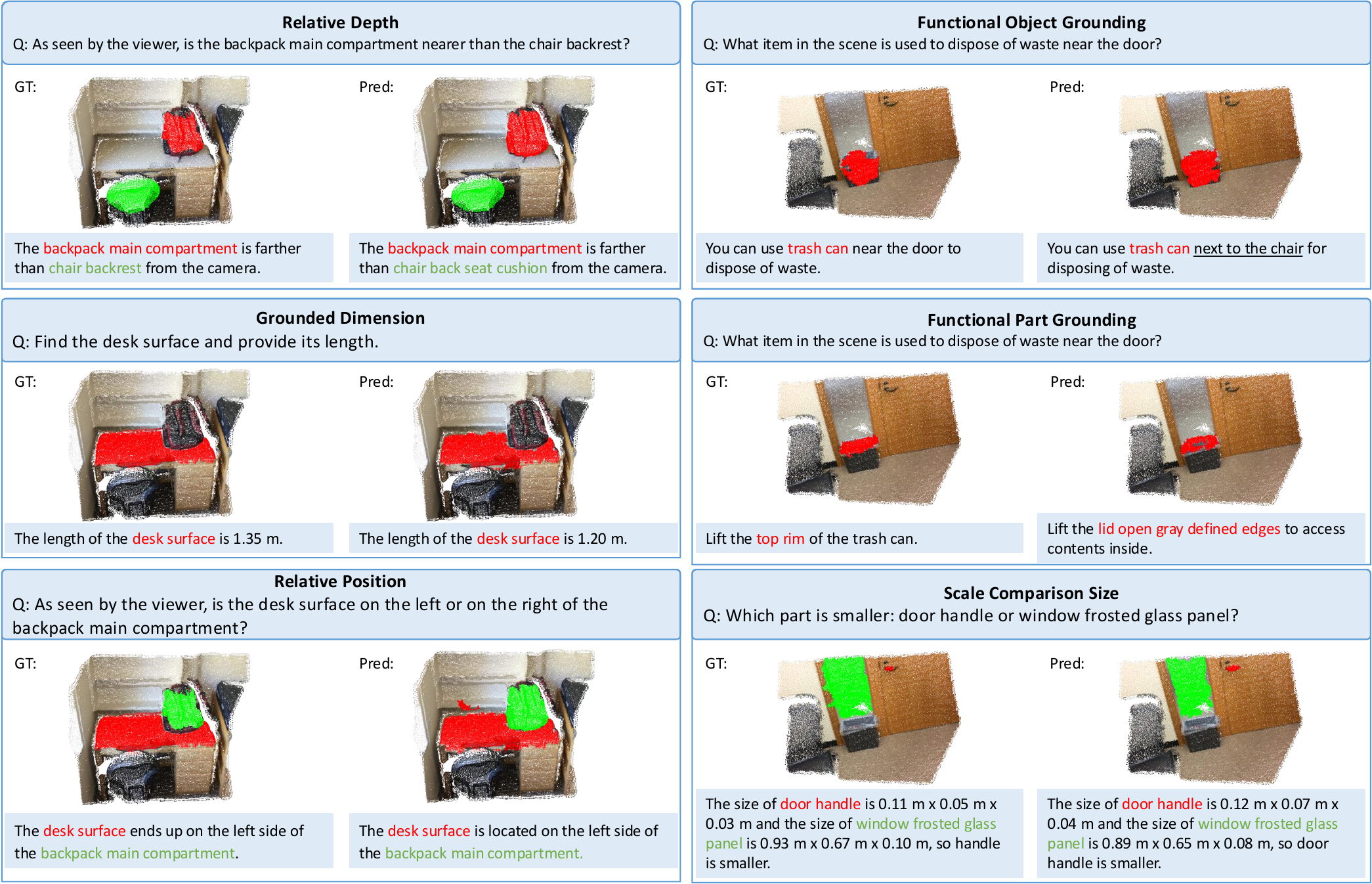}
    \caption{Qualitative results of Ground3D-LMM across six representative object- and part-level sub-tasks. (Zoom in for better visualization.)}
    \label{fig:qualitative}
\end{figure}

\subsection{Ablation Study}
We ablate three design choices on Ground3D-ScanNet object-level results (Table~\ref{tab:ablation}): \emph{(a)} reducing training data to 25\%/50\% degrades all metrics, confirming the full dataset is necessary; \emph{(b)} text-only supervision eliminates mask prediction and weakens metric quality, showing joint mask and metric supervision are mutually beneficial; \emph{(c)} randomly zeroing up to 50\% of ground-truth masks causes only modest mIoU drops, indicating robustness to noisy segmentation supervision.

\input{tables/table_ablation}

\subsection{Qualitative Results}
In Fig.~\ref{fig:qualitative}, we present qualitative results of Ground3D-LMM across six sub-tasks. 
Our model performs well on tasks involving relative position, scale comparison, and metric estimation. For example, in the Scale Comparison and Grounded Dimension task, the prediction error is within 0.05 m.
Moreover, from functional object grounding to functional part grounding, the model accurately localizes and describes both objects and their corresponding parts within the same scene. The consistent performance at both object and part levels highlights the effectiveness of our method and the validity of the proposed dataset.


%% file: tables/table_ourdata_split.tex
\begin{table}[t]
	\caption{Object segmentation results (mIoU) on Ground3D-ScanNet and Ground3D-ScanNet++. Our Ground3D-LMM achieves superior performance across both 3D-only and multimodal (3D+2D) settings.}
	\centering
	\footnotesize
	\setlength{\tabcolsep}{5pt}
	\renewcommand{\arraystretch}{1.2}
	\resizebox{\textwidth}{!}{%
		\begin{tabular}{l c | c c c c c c c c}
			\toprule
			\textbf{Method} & \textbf{Modality} & \textbf{Overall} & \textbf{Dist.} & \makecell[c]{\textbf{Func.} \\ \textbf{Obj.}} & \makecell[c]{\textbf{Gnd.} \\ \textbf{Mea.}} & \makecell[c]{\textbf{Depth} \\ \textbf{Rel.}} & \makecell[c]{\textbf{Spa.} \\ \textbf{Rel.}} & \makecell[c]{\textbf{Size} \\ \textbf{Comp.}} & \makecell[c]{\textbf{Metric} \\ \textbf{Est.}} \\
			\midrule
			\midrule
			\multicolumn{10}{l}{\textit{Ground3D-ScanNet}} \\
			Image baseline & 2D & 32.52 & 31.46 & 36.67 & 37.51 & 32.08 & 28.11 & 25.66 & 36.18 \\
            Reason3D~\cite{huang2024reason3d} & 3D & 9.31 & -- & 6.09 & 12.07 & -- & -- & -- & 9.76 \\
			UniSeg3D~\cite{xu2024unified}       & 3D & 22.97 & 23.64 & 24.00 & 23.87 & 22.56 & 21.71 & 21.13 & 23.87 \\
			\rowcolor{hl!30} Ground3D-LMM & 3D & 37.40 & 37.79 & 37.64 & 39.10 & 35.79 & 37.42 & 34.89 & 39.14 \\
			\rowcolor{hl} Ground3D-LMM & 3D+2D & \textbf{42.22} & \textbf{41.37} & \textbf{43.73} & \textbf{44.62} & \textbf{42.59} & \textbf{40.42} & \textbf{38.56} & \textbf{44.28} \\
			\midrule
			\multicolumn{10}{l}{\textit{Ground3D-ScanNet++}} \\
			Image baseline & 2D & 27.80 & 27.97 & 23.56 & 30.08 & 27.70 & 28.21 & 28.49 & 28.60 \\
            Reason3D~\cite{huang2024reason3d} & 3D & 2.89 & -- & 1.34 & 3.95 & -- & -- & -- & 3.39 \\
			UniSeg3D ~\cite{xu2024unified}      & 3D & 10.84 & 11.84 & 10.86 & 10.83 & 11.16 & 10.95 & 9.39  & 10.82 \\
			\rowcolor{hl!30} Ground3D-LMM & 3D & 24.56 & 24.70 & 18.56 & 26.04 & 24.11 & 26.86 & 25.85 & 25.79 \\
			\rowcolor{hl} Ground3D-LMM & 3D+2D & \textbf{30.42} & \textbf{30.97} & \textbf{20.27} & \textbf{33.08} & \textbf{31.60} & \textbf{34.55} & \textbf{32.44} & \textbf{30.01} \\
			\bottomrule
		\end{tabular}%
	}
	\label{tab:object_segmentation}
\end{table}

\begin{table}[t]
	\caption{Part segmentation results (mIoU) on Ground3D-ScanNet and Ground3D-ScanNet++. Our method demonstrates strong generalization in fine-grained part reasoning, especially when integrated with 2D visual cues.}
	\centering
	\footnotesize
	\setlength{\tabcolsep}{4pt}
	\renewcommand{\arraystretch}{1.2}
	\resizebox{\textwidth}{!}{%
		\begin{tabular}{l c | c c c c c c c c c}
			\toprule
			\textbf{Method} & \textbf{Modality} & \textbf{Overall} & \textbf{Dist.} & \textbf{Exist.} & \makecell[c]{\textbf{Func.} \\ \textbf{Part}} & \makecell[c]{\textbf{Gnd.} \\ \textbf{Mea.}} & \makecell[c]{\textbf{Depth} \\ \textbf{Rel.}} & \makecell[c]{\textbf{Spa.} \\ \textbf{Rel.}} & \makecell[c]{\textbf{Size} \\ \textbf{Comp.}} & \makecell[c]{\textbf{Metric} \\ \textbf{Est.}} \\
			\midrule
			\midrule
			\multicolumn{11}{l}{\textit{Ground3D-ScanNet}} \\
			Image baseline & 2D & 21.52 & 20.70 & 22.94 & 24.73 & 22.82 & 20.21 & 15.35 & 20.14 & 25.28 \\
            Reason3D~\cite{huang2024reason3d} & 3D & 8.19 & -- & -- & 7.04 & 9.35 & -- & -- & -- & 8.18 \\
			UniSeg3D~\cite{xu2024unified}       & 3D & 11.28 & 11.71 & 9.70  & 10.14 & 12.25 & 11.69 & 10.77 & 11.70 & 12.25 \\
			\rowcolor{hl!30} Ground3D-LMM & 3D & 30.06 & 30.47 & 32.20 & 27.27 & 30.83 & 30.36 & 28.15 & 30.31 & 30.89 \\
			\rowcolor{hl} Ground3D-LMM & 3D+2D & \textbf{36.57} & \textbf{37.33} & \textbf{38.21} & \textbf{31.70} & \textbf{38.24} & \textbf{37.01} & \textbf{34.79} & \textbf{37.01} & \textbf{38.28} \\
			\midrule
			\multicolumn{11}{l}{\textit{Ground3D-ScanNet++}} \\
			Image baseline & 2D & 25.86 & 25.66 & 29.59 & 26.86 & 27.00 & 24.39 & 22.33 & 24.90 & 26.13 \\
            Reason3D~\cite{huang2024reason3d} & 3D & 4.91 & -- & -- & 4.04 & 4.92 & -- & -- & -- & 5.77 \\
			UniSeg3D  ~\cite{xu2024unified}     & 3D & 8.03  & 8.62  & 10.20 & 7.27  & 7.18  & 7.75  & 7.75  & 8.24  & 7.28  \\
			\rowcolor{hl!30} Ground3D-LMM & 3D & 29.80 & 30.69 & 34.07 & 27.02 & 28.77 & 29.81 & 28.93 & 29.62 & 29.49 \\
			\rowcolor{hl} Ground3D-LMM & 3D+2D & \textbf{39.07} & \textbf{40.14} & \textbf{43.48} & \textbf{35.39} & \textbf{39.37} & \textbf{39.03} & \textbf{37.56} & \textbf{38.82} & \textbf{38.80} \\
			\bottomrule
		\end{tabular}%
	}
	\label{tab:part_segmentation}
\end{table}

\begin{table}[t]
\caption{Text and metric evaluation on Ground3D-ScanNet and Ground3D-ScanNet++ evaluation subsets. Higher is better except Mean APE (lower is better).}
\centering
\footnotesize
\renewcommand{\arraystretch}{1.2}
\setlength{\tabcolsep}{4pt}

\resizebox{\textwidth}{!}{%
\begin{tabular}{@{} l c
  @{\hspace{10pt}}
  S[table-format=1.2] S[table-format=1.2] S[table-format=3.2] S[table-format=2.2]
  @{\hspace{14pt}}
  S[table-format=1.2] S[table-format=1.2] S[table-format=3.2] S[table-format=2.2]
@{}}
\toprule
\multirow{2}{*}{\textbf{Method}} &
\multirow{2}{*}{\textbf{Modality}} &
\multicolumn{4}{c}{\textit{Ground3D ScanNet Subset}} &
\multicolumn{4}{c}{\textit{Ground3D ScanNet++ Subset}} \\
\cmidrule(lr){3-6}\cmidrule(lr){7-10}
& &
\multicolumn{1}{c}{\textbf{Halluc.}} &
\multicolumn{1}{c}{\textbf{Comp.}} &
\multicolumn{1}{c}{\makecell[c]{\textbf{Mean}\\\textbf{APE}}} &
\multicolumn{1}{c}{\makecell[c]{\boldmath$\boldsymbol{\delta}$\\\textbf{Succ. Rate}}} &
\multicolumn{1}{c}{\textbf{Halluc.}} &
\multicolumn{1}{c}{\textbf{Comp.}} &
\multicolumn{1}{c}{\makecell[c]{\textbf{Mean}\\\textbf{APE}}} &
\multicolumn{1}{c}{\makecell[c]{\boldmath$\boldsymbol{\delta}$\\\textbf{Succ. Rate}}} \\
\midrule

Image baseline              & 2D    & 8.06 & 7.37 & 166.91 & 25.05 & 8.29 & 7.55 & 120.22 & 24.17 \\
SD-VLM~\cite{chen2025sd}                      & 2D    & 8.01 & 8.27 & 231.03 & 20.12 & 8.19 & 7.64 & 183.93 & 15.46 \\
\rowcolor{hl!30} Ground3D-LMM & 3D    & 9.30 & 8.38 &  79.85 & 34.78 & 9.30 & 7.97 &  84.15 & 27.49 \\
\rowcolor{hl}    Ground3D-LMM & 3D+2D & 9.16 & 7.99 &  74.03 & 43.55 & 9.05 & 7.39 &  88.20 & 33.93 \\

\bottomrule
\end{tabular}%
}
\label{tab:text_metric_eval}
\end{table}

%% file: tables/table_reason3d.tex
\begin{table}[!t]
	\caption{\textbf{Segmentation results on Reason3D.} The evaluation metrics include accuracy at IoU thresholds 0.25, 0.50, and mean IoU.}
	\label{tab:reason3d}
	\centering

	\renewcommand{\arraystretch}{1.2}
	\resizebox{0.8\linewidth}{!}{
		\begin{tabular}{l l c ccc}
			\toprule
			{\textbf{Methods}} & 
			{\textbf{Venue}} &
			{\textbf{Modality}} & \textbf{Acc@0.25} & \textbf{Acc@0.50} & \textbf{mIoU}  \\
			\midrule
			3D-STMN~\cite{3dstmn} & AAAI'24 & 3D & 25.43 & 17.78 & 18.23  \\
			Intent3D~\cite{intent3d} & ICLR'25 & 3D & 20.57 & 19.46 & -  \\
			
			Reason3D~\cite{huang2024reason3d} & 3DV'25 & 3D & 43.21 & 32.10 & 31.20 \\
			\rowcolor{hl}Ground3D-LMM (Ours) &-&3D&\textbf{50.32}&\textbf{37.01}&\textbf{36.35}\\
			\midrule
			MLLM-For3D~\cite{huang2025mllmfor3d}+ LISA-7B  & NeurIPS'25 & 3D+2D & 45.53 & 39.54 & 31.94   \\
			MLLM-For3D~\cite{huang2025mllmfor3d}+ VideoLISA & NeurIPS'25 & 3D+2D & {48.45} & {41.02} & {39.82}  \\
			\rowcolor{hl}Ground3D-LMM (Ours)&-&3D+2D&\textbf{57.79}&\textbf{41.23} & \textbf{41.29}\\
			\bottomrule
		\end{tabular}
	}
\end{table}

%% file: tables/tabel_scanrefer.tex
\begin{table}[!t]
	\caption{\textbf{Evaluation results on ScanRefer dataset.} The evaluation metrics include accuracy at IoU thresholds 0.25, 0.50, and mean IoU.}
	\label{tab:scanrefer}
	\centering
	\small
	\renewcommand{\arraystretch}{1.2}
	\resizebox{0.8\linewidth}{!}{
		\begin{tabular}{l l c c c c}
			\toprule
			\textbf{Methods} & \textbf{Venue} & \textbf{Modality}& \textbf{Acc@0.25} $\uparrow$ & \textbf{Acc@0.50} $\uparrow$ & \textbf{mIoU} $\uparrow$ \\
			\midrule
			ScanRefer~\cite{chen2020scanrefer} & ECCV'20 & 3D & 10.51 & 6.20  & - \\
			TGNN\ ~\cite{tgnn}                   & AAAI'21 &3D& 11.64 & 9.51  & 8.13 \\
			InstanceRefer~\cite{Yuan_2021_ICCV}    & ICCV'21 &3D& 13.92 & 11.47 & - \\
			BUTD-DETR~\cite{butd_detr}         & ECCV'22 &3D& 11.99 & 8.95  & - \\
			M3DRef-CLIP~\cite{zhang2023multi3drefer}  & ICCV'23& 3D & 18.3  & 14.8  & 10.29 \\
			EDA~\cite{wu2023eda}               & CVPR'23 & 3D &26.50 & 21.20 & - \\
			Reason3D~\cite{huang2024reason3d}  & 3DV'25& 3D & 17.64 & 13.11 & 13.05 \\
			IntentNet~\cite{intent3d}    & ICLR'25 & 3D& 28.12 & 22.63 & 18.92 \\
            UniSeg3D~\cite{xu2024unified} &NeurIPS'24& 3D & -&-&29.10\\
				\rowcolor{hl}Ground3D-LMM (Ours) &-&3D& \textbf{55.73}&\textbf{32.71}&\textbf{38.72}\\
			\midrule
		   MLLM-For3D~\cite{huang2025mllmfor3d}+ LISA-7B & NeurIPS'25& 3D+2D& 31.88 & 29.90 & 28.10 \\
			MLLM-For3D~\cite{huang2025mllmfor3d}+VideoLISA  & NeurIPS'25& 3D+2D & {33.12} & {31.21} & {30.45} \\
				\rowcolor{hl}Ground3D-LMM (Ours) &-&3D+2D& \textbf{59.98}&\textbf{36.37}&\textbf{41.30}\\
			\bottomrule
		\end{tabular}
	}
\end{table}

%% file: tables/table_ablation.tex
\begin{table}[t]
	\caption{\textbf{Ablation study on the object-level task} (Ground3D-ScanNet). We ablate \emph{(a)} dataset scale, \emph{(b)} joint mask + metric supervision, and \emph{(c)} segmentation supervision quality. Higher is better except APE (lower is better).}
	\label{tab:ablation}
	\centering
	\footnotesize
	\renewcommand{\arraystretch}{1.2}
	\setlength{\tabcolsep}{6pt}
	\resizebox{0.85\linewidth}{!}{%
		\begin{tabular}{l ccccc}
			\toprule
			\textbf{Setting} &
			\textbf{mIoU}$\uparrow$ &
			\textbf{APE}$\downarrow$ &
			\boldmath$\delta\,\uparrow$ &
			\textbf{Halluc.}$\uparrow$ &
			\textbf{Comp.}$\uparrow$ \\
			\midrule
			\rowcolor{hl} Ground3D-LMM (Ours, 3D+2D) & 42.22 & 76.07 & 46.34 & 9.16 & 7.77 \\
			\midrule
			\emph{(a)} Ours w.\ 25\% data        & 36.93 & 99.23 & 40.26 & 8.86 & 7.03 \\
			\emph{(a)} Ours w.\ 50\% data        & 39.64 & 81.89 & 40.84 & 8.93 & 7.26 \\
			\midrule
			\emph{(b)} Ours w.\ text-only        & --    & 77.41 & 45.62 & 9.02 & 7.35 \\
			\midrule
			\emph{(c)} Ours w.\ 25\% noise mask  & 41.43 & 77.97 & 44.09 & 8.91 & 7.32 \\
			\emph{(c)} Ours w.\ 50\% noise mask  & 41.06 & 79.60 & 43.84 & 9.01 & 7.45 \\
			\bottomrule
		\end{tabular}%
	}
\end{table}

%% file: sections/6_conclusion.tex
\section{Conclusion}\label{sec:conclusion}
We introduced Ground3D-LMM, a unified point-cloud multimodal model that produces verifiable 3D conversational outputs by pairing natural-language responses with point-level masks, and supports metric measurements in consistent real-world units. We also proposed the 3D Grounded Measurement task and the Ground3D dataset, which jointly benchmark object- and part-level grounding, multi-object spatial relations, and multi-turn dialogue. Experiments on Ground3D, Reason3D, and ScanRefer demonstrate strong gains over open-vocabulary segmentation and 3D reasoning baselines, and highlight the benefit of incorporating an RGB view when available.

%% file: sections/7_acknowledgment.tex
\section{Acknowledgment}
The computations were enabled by resources provided by LUMI hosted by CSC (Finland) and LUMI consortium, and by Berzelius resource provided by the Knut and Alice Wallenberg Foundation at the NSC.

%% file: appendix/A_semantic_novelty.tex
\section{Semantic Novelty Beyond the Native Taxonomy}
\label{sec:semantic_novelty}

A central goal of our Ground3D dataset is to introduce new kinds of labels (i.e., semantically novel object/part concepts), rather than simply producing more labels due to different frames or viewpoints. Since raw label counts are highly sensitive to sampling and repetition, we instead evaluate novelty in a meaning space: we embed label strings using the OpenCLIP ViT-B/32 text encoder pretrained on LAION-2B \cite{radford2021clip,ilharco2021openclip,schuhmann2022laion5b}, producing 512-dimensional L2-normalized text embeddings.

\begin{figure}[ht!]
    \centering
    \includegraphics[width=0.6\linewidth]{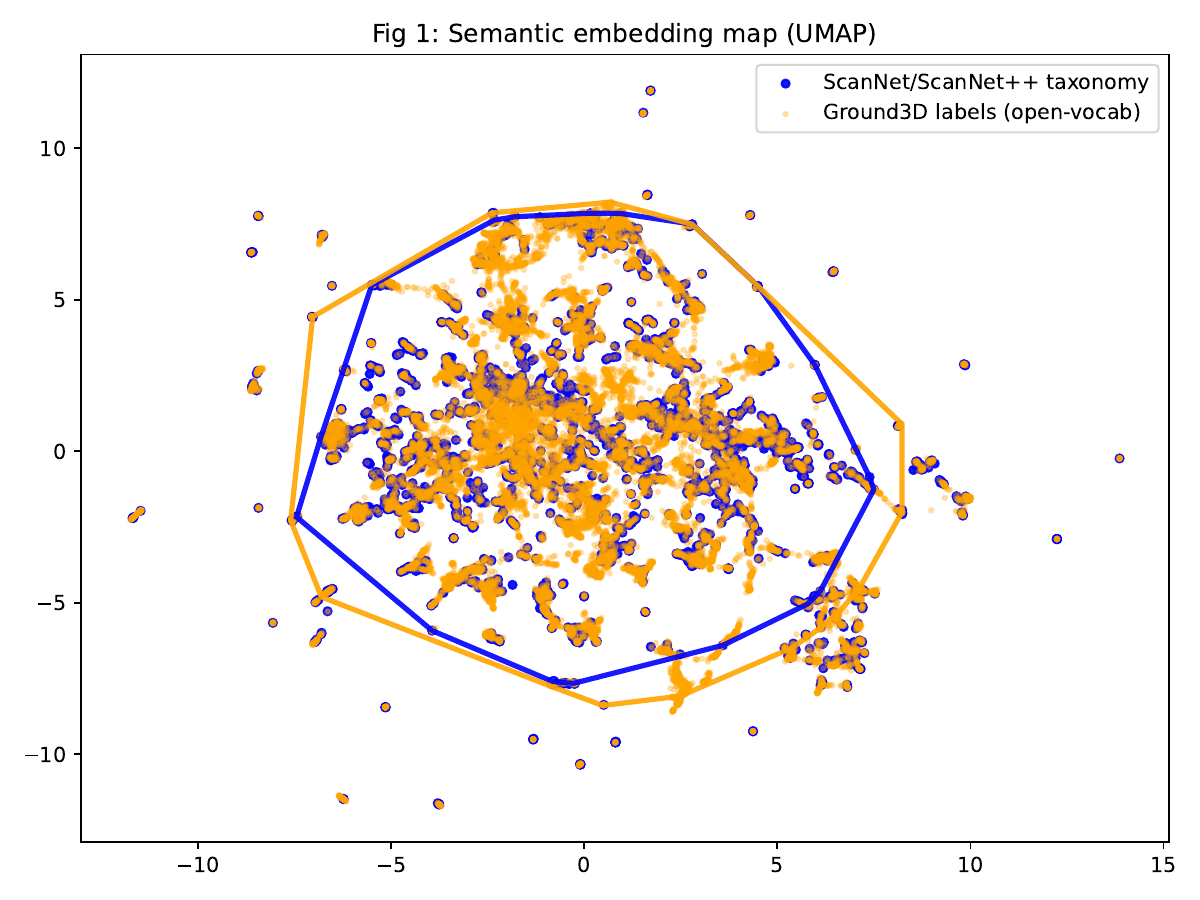}
    \caption{\textbf{Semantic embedding map (UMAP).}
    CLIP text embeddings of ScanNet/ScanNet++ taxonomy labels (blue) and Ground3D labels (orange) projected to 2D using UMAP.}
    \label{fig:umap_semantic_map}
\end{figure}

\paragraph{Qualitative evidence: semantic space expansion.}
Figure~\ref{fig:umap_semantic_map} visualizes the semantic coverage of Ground3D dataset relative to the native ScanNet/ScanNet++ taxonomy.
We compute CLIP text embeddings for all taxonomy labels (ScanNet/ScanNet++) and all Ground3D labels (3,419 baseline taxonomy labels and 25,569 Ground3D labels after deduplication), and project them into 2D using Uniform Manifold Approximation and Projection (UMAP) \cite{mcinnes2018umap} with cosine distance (n\_neighbors=30, min\_dist=0.1). Cosine similarity is used as the distance metric, consistent with standard practice for normalized text embeddings.
As shown, taxonomy labels form a relatively compact semantic core, reflecting a constrained predefined vocabulary.
In contrast, Ground3D labels spread into a broader region of the embedding space and extend beyond the taxonomy envelope, indicating additional semantic neighborhoods not represented in the original taxonomy.
To avoid the hull being dominated by a small number of projection outliers, we plot a trimmed convex hull that encloses the inner 90\% of points (by distance to the centroid), which provides a more stable visualization.

\begin{figure}[ht!]
    \centering
    \includegraphics[width=0.6\linewidth]{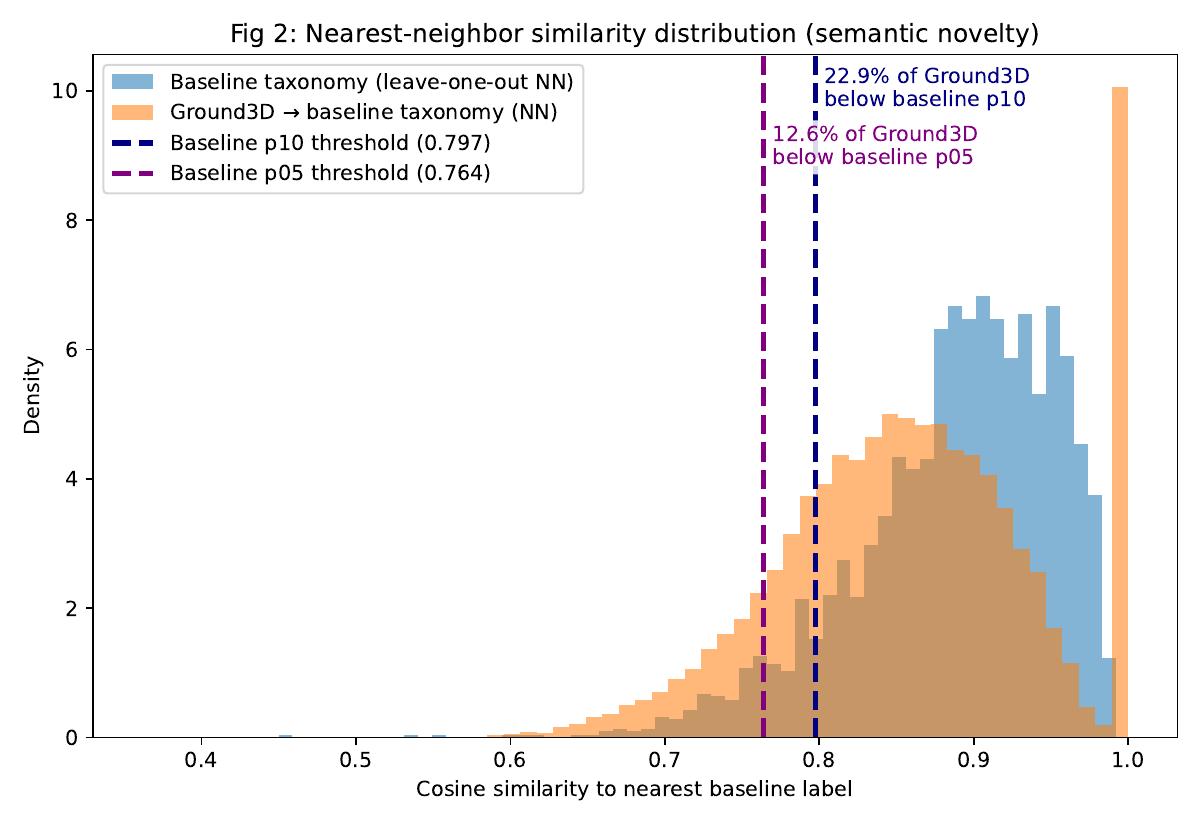}
    \caption{\textbf{Nearest-neighbor similarity distribution (semantic novelty).}
    \emph{Blue:} leave-one-out NN cosine similarity within the ScanNet/ScanNet++ taxonomy.
    \emph{Orange:} NN cosine similarity from each Ground3D label to its nearest taxonomy label.
    Dashed vertical lines mark baseline-driven novelty thresholds.
    A substantial fraction of Ground3D labels lies below these thresholds, indicating semantic expansion beyond the native taxonomy.}
    \label{fig:nn_similarity_hist}
\end{figure}

\paragraph{Quantitative evidence: nearest-neighbor similarity to the taxonomy.}
While the UMAP visualization provides intuitive evidence of expansion, we further quantify novelty using nearest-neighbor (NN) cosine similarity in the original CLIP embedding space (Figure~\ref{fig:nn_similarity_hist}).
First, we establish what normal semantic closeness looks like within the baseline taxonomy (native ScanNet/ScanNet++) by computing a leave-one-out NN similarity for each taxonomy label with the nearest other taxonomy label.
This yields a high-cohesion baseline distribution (mean $0.888$, with lower-tail thresholds $p10=0.797$ and $p05=0.764$), indicating that even taxonomy labels that are relatively distinct from each other typically remain above $\sim 0.76$ similarity.

Next, for each Ground3D dataset label we compute its cosine similarity to the nearest baseline taxonomy label.
This distribution is shifted left (mean $0.858$), with a substantially heavier low-similarity tail.
Crucially, $22.9\%$ of Ground3D labels fall below the taxonomy $p10$ threshold ($0.797$), and $12.6\%$ fall below the taxonomy $p05$ threshold ($0.764$).
In other words, approximately ~25\% of Ground3D labels fall outside the typical semantic similarity range observed within the baseline taxonomy, indicating the introduction of genuinely novel concepts rather than superficial lexical variations.

Together, Figure~\ref{fig:umap_semantic_map} and Figure~\ref{fig:nn_similarity_hist} provide complementary evidence that Ground3D dataset expands the semantic support of the ScanNet/ScanNet++ taxonomy, demonstrating meaningful semantic novelty rather than simply producing more labels.

%% file: appendix/B_qualitative_results.tex
\section{Qualitative Results}

\begin{figure}[ht!]
    \centering
    \includegraphics[width=0.9\linewidth]{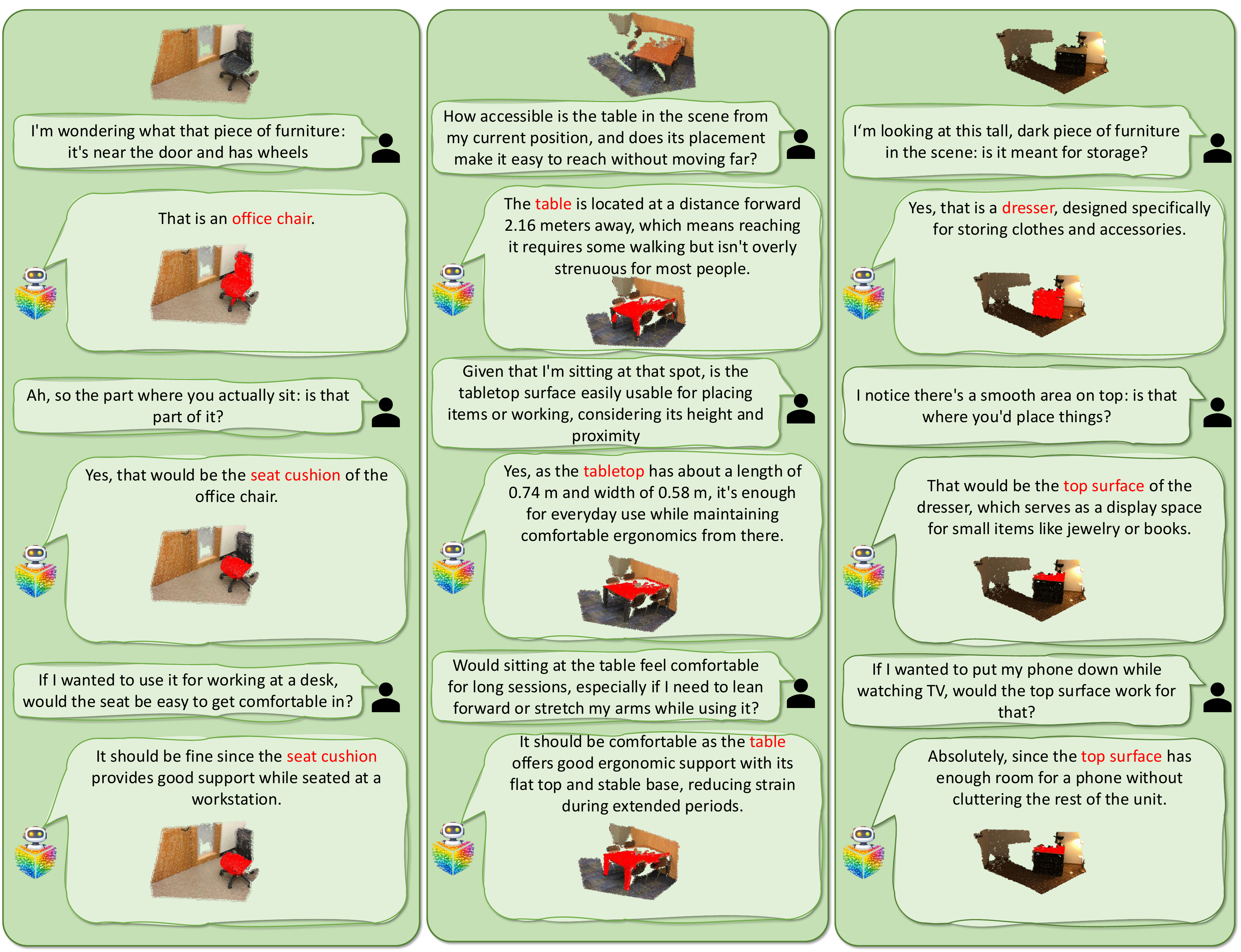}
    \caption{Qualitative results of Ground3D-LMM on 3D multi-turn conversation. (Zoom in for better visualization.)}
    \label{fig:multi_turn_visual}
\end{figure}

\begin{figure}[ht!]
    \centering
    \includegraphics[width=\linewidth]{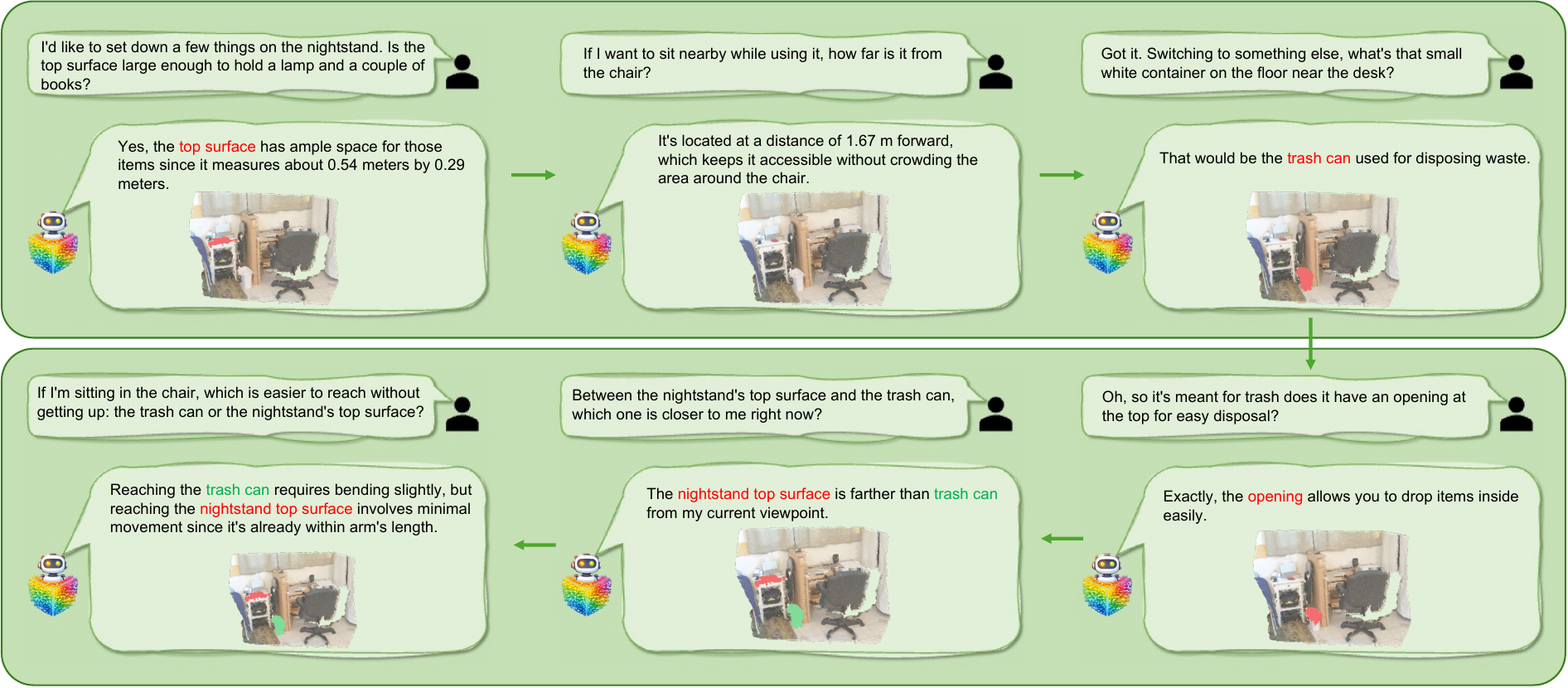}
    \caption{Additional qualitative results of Ground3D-LMM on a six-turn 3D conversation. Note that multi-turn dialogues of this form are not present in our training data; the model generalizes to them at inference. (Zoom in for better visualization.)}
    \label{fig:multi_turn_visual2}
\end{figure}

\begin{figure}[ht!]
    \centering
    \includegraphics[width=0.9\linewidth]{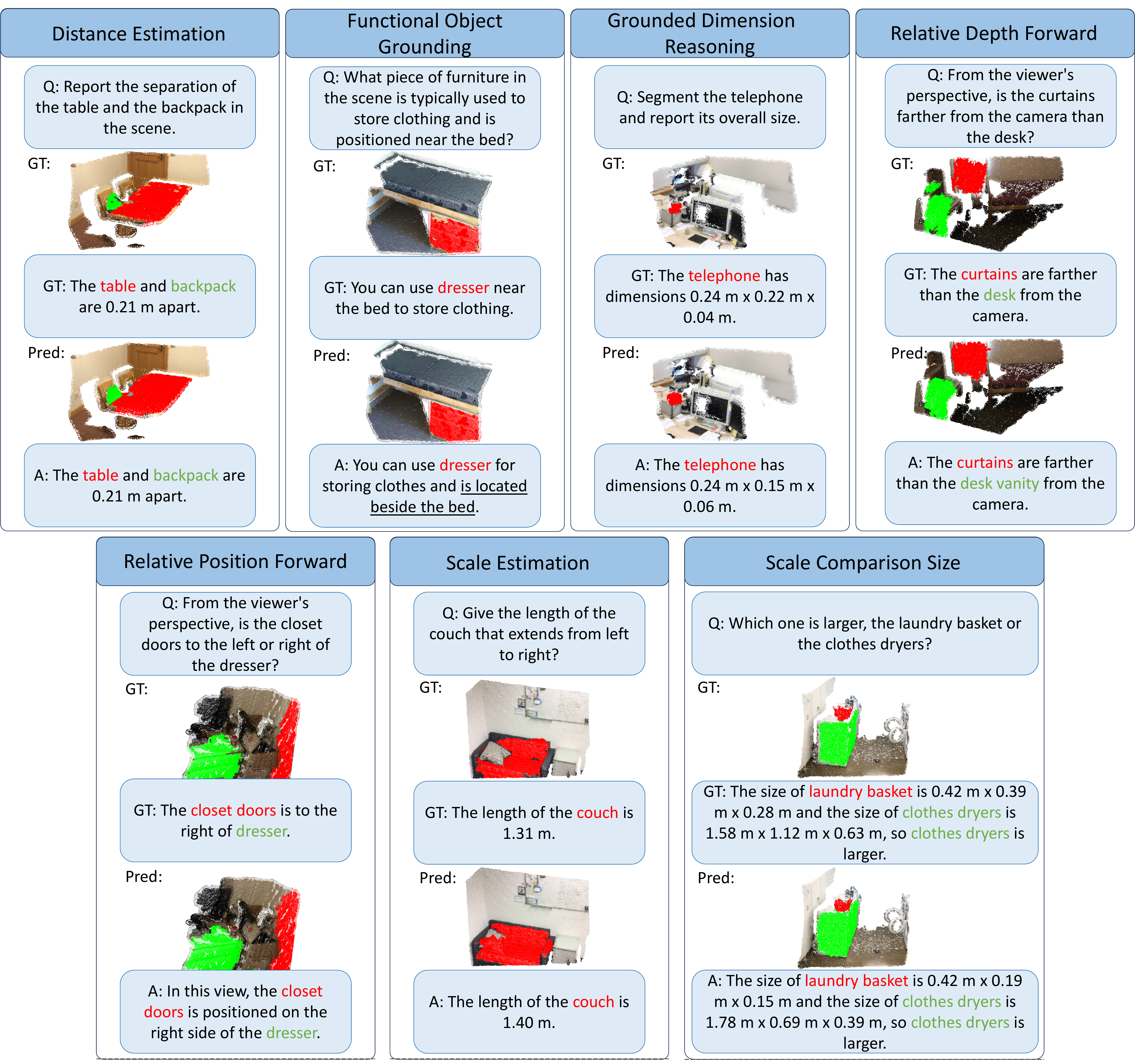}
    \caption{Qualitative results of Ground3D-LMM on all object-level tasks. (Zoom in for better visualization.)}
    \label{fig:object_level_visual}
\end{figure}

\begin{figure}[p]
    \centering
    \includegraphics[height=0.9\textheight]{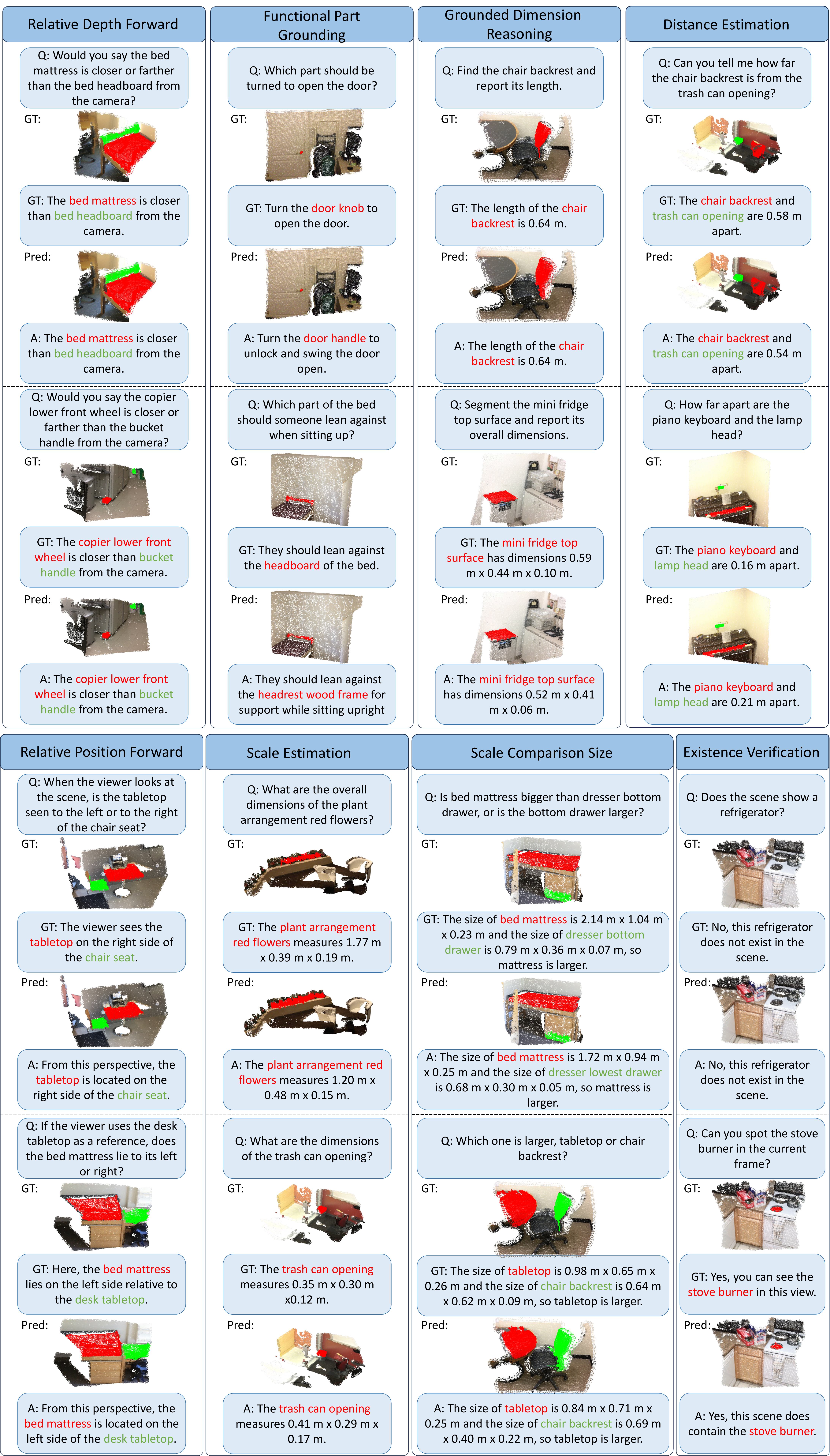}
    \caption{Qualitative results of Ground3D-LMM on all part-level tasks. (Zoom in for better visualization.)}
    \label{fig:part_level_visual}
\end{figure}

In this section, we provide more comprehensive qualitative results with additional examples from our dataset that highlight the performance of our method. Fig.~\ref{fig:multi_turn_visual} illustrates qualitative results on 3D multi-turn conversations. Fig.~\ref{fig:multi_turn_visual2} further shows a longer six-turn conversation that grounds multiple referents in one scene, carrying context across turns and reasoning from different viewpoints (proximity to the camera vs.\ reachability when seated). Moreover, Fig.~\ref{fig:part_level_visual} presents results across all part-level tasks, and Fig.~\ref{fig:object_level_visual} shows performance on various object-level tasks.

%% file: appendix/C_additional_quantitative_results.tex
\section{Additional Quantitative Results}

\subsection{Grounded Measurement}
Table~\ref{tab:grounded_measurement} reports the Grounded Measurement subtask, which consolidates grounding (mIoU), metric accuracy (APE, $\delta$), and text quality (Hallucination, Completeness) into a single view. The Grounded-Measurement Success Rate (GM-$\delta$) is defined in the main paper. Averaged across Ground3D-ScanNet/ScanNet++ and both object and part levels, Ground3D-LMM attains a GM-$\delta$ of $42.69$, confirming that our predictions are simultaneously well-grounded and metrically accurate rather than excelling on only one axis.

\begin{table}[t]
	\caption{\textbf{Grounded Measurement subtask}, averaged across Ground3D-ScanNet/ScanNet++ and object/part levels. Higher is better except APE (lower is better).}
	\label{tab:grounded_measurement}
	\centering
	\footnotesize
	\renewcommand{\arraystretch}{1.2}
	\setlength{\tabcolsep}{6pt}
	\resizebox{\linewidth}{!}{%
		\begin{tabular}{l cccccc}
			\toprule
			\textbf{Method} & \textbf{mIoU}$\uparrow$ & \textbf{APE}$\downarrow$ & \boldmath$\delta\,\uparrow$ & \textbf{Halluc.}$\uparrow$ & \textbf{Comp.}$\uparrow$ & \textbf{GM-}\boldmath$\delta\,\uparrow$ \\
			\midrule
			\rowcolor{hl} Ground3D-LMM (Ours, 3D+2D) & 38.83 & 72.92 & 44.28 & 9.76 & 9.49 & 42.69 \\
			\bottomrule
		\end{tabular}%
	}
\end{table}

\subsection{Generality Beyond ScanNet}
To assess generalization beyond ScanNet-based benchmarks, we evaluate on CA-VQA~\cite{daxberger2025mmspatial} in Table~\ref{tab:cavqa}. Ground3D-LMM achieves an average score of $48.1$, surpassing GPT-4o~\cite{openai2024gpt4}, SpatialRGPT-8B~\cite{cheng2024spatialrgpt}, and MM1.5-3B~\cite{zhang2024mm15}, and is on par with MM-Spatial~\cite{daxberger2025mmspatial} (comparing against variants that do not use chain-of-thought or multi-view inputs, as these are beyond our scope). Notably, our model outperforms all variants on the binary, counting, multi-choice, and object-size metrics, confirming that its grounded metric reasoning transfers well to new data distributions.

\begin{table*}[t]
	\caption{\textbf{Results on CA-VQA}, evaluating generality beyond ScanNet. Ground3D-LMM surpasses GPT-4o, SpatialRGPT-8B, and MM1.5-3B on the average score. All values are higher-is-better.}
	\label{tab:cavqa}
	\centering
	\footnotesize
	\renewcommand{\arraystretch}{1.2}
	\setlength{\tabcolsep}{5pt}
	\resizebox{\textwidth}{!}{%
		\begin{tabular}{l ccccccccc}
			\toprule
			\textbf{Method} & \textbf{Binary} & \textbf{Count} & \makecell[c]{\textbf{2D}\\\textbf{Gnd}} & \makecell[c]{\textbf{3D}\\\textbf{Gnd}} & \textbf{Multi-c} & \makecell[c]{\textbf{Ego-}\\\textbf{Dist}} & \makecell[c]{\textbf{Obj-}\\\textbf{Dist}} & \makecell[c]{\textbf{Obj-}\\\textbf{Size}} & \textbf{Avg} \\
			\midrule
			GPT-4o~\cite{openai2024gpt4} & 44.2 & 69.0 &  0.0 &  0.0 & 36.6 & 11.7 & 10.0 & 11.0 & 22.8 \\
			SpatialRGPT-8B~\cite{cheng2024spatialrgpt} & 53.6 & 68.8 &  5.5 &  0.0 & 37.2 & 10.5 &  8.7 &  7.0 & 23.9 \\
			MM1.5-3B~\cite{zhang2024mm15} & 59.1 &  9.1 & 32.6 &  0.0 & 38.6 &  0.6 &  2.2 &  3.4 & 18.2 \\
			\rowcolor{hl} Ground3D-LMM (Ours, 3D+2D) & \textbf{81.2} & \textbf{82.8} & \textbf{41.2} & \textbf{11.5} & \textbf{85.7} & \textbf{37.8} & \textbf{14.2} & \textbf{30.0} & \textbf{48.1} \\
			\bottomrule
		\end{tabular}%
	}
\end{table*}

%% file: appendix/C_data_prompts.tex
\lstdefinestyle{promptstyle}{
  basicstyle=\ttfamily\scriptsize,
  breaklines=true,
  breakatwhitespace=false,
  columns=fullflexible,
  keepspaces=true,
  showstringspaces=false,
  frame=single,
  framesep=4pt,
  xleftmargin=4pt,
  xrightmargin=4pt,
}

\section{Data Generation Prompts and Settings}
\label{app:prompts}

For full reproducibility we describe the prompts used by our annotation
pipeline (Section~3 of the main paper). To keep this section compact, we
factor out the conventions shared by all prompts (Section~\ref{app:shared}),
give one prompt verbatim (Section~\ref{app:verbatim}), and tabulate only the
per-task differences (Section~\ref{app:deltas}). Each synthesis prompt has an
object-level and a part-level instantiation that are \emph{identical}
except that the referent is the whole object (\texttt{object\_name}) or
one of its salient parts (\texttt{part\_label}); we therefore describe
them jointly.

\subsection{Shared Generation Protocol}
\label{app:shared}


\paragraph{Message format.} Each prompt is a single user message
composed of (a) a short task description, (b) a rules-and-schema block
containing one worked example, and (c) a compact JSON context holding
the per-referent metrics or pairwise relations computed in Stage~4.
The annotated scene image is attached so the
model can resolve which labeled instance is intended.

\paragraph{Shared constraints.} Every synthesis prompt enforces the same
output discipline: return \emph{only} valid JSON (no Markdown fences,
double-quoted keys, a fixed \texttt{results} schema); use \emph{only}
the numeric values supplied in the context and never invent numbers;
report lengths in meters to two decimals; and wrap each referent name in
\texttt{<p>...</p>} immediately followed by a \texttt{<SEG>} token, in
answers only (never in questions). When several instances of the same
category are visible, the question disambiguates the target using
\emph{only} natural spatial cues (left/right/middle, nearest/farthest,
or relations such as "near the door"); numeric ids and coordinates are
never surfaced in text. Phrasing is diversified by requiring varied
question openers.

\paragraph{Decoding.} All generation uses nucleus sampling with
\texttt{max\_new\_tokens}~$=1524$. Temperature ranges from $0.1$ for the
templated single-referent metric prompts to $0.4$ for the more
open-ended functional and object-level prompts, with \texttt{top\_p} in
$[0.9, 0.95]$; the higher-temperature runs additionally use
\texttt{top\_k}~$=50$. Outputs are
parsed by extracting the first balanced JSON object and normalizing
smart quotes and trailing commas before \texttt{json.loads}; entries
that fail to parse are discarded.

\subsection{Per-Task Differences}
\label{app:deltas}

Table~\ref{tab:deltas} summarizes the eight subtasks: the Stage-4 context fields each prompt reads, how many QA pairs it produces, and any task-specific notes. The \emph{Spatial Relations} and \emph{Depth Relations} prompts can also generate a reverse variant that swaps the referent order and flips the direction label (left$\leftrightarrow$right, closer$\leftrightarrow$farther); everything else stays the same. To keep the dataset balanced across subtasks, we cap the number of QA pairs retained per frame during filtering.

\subsection{Representative Prompt}
\label{app:verbatim}

Listing~\ref{lst:scale} shows one full prompt used to generate the QA data for
the \emph{Metric Estimation} subtask (the task message followed by its
rules-and-schema block). The prompts for all other subtasks use the same
template and differ only as summarized in Table~\ref{tab:deltas}.
\clearpage
\begin{lstlisting}[style=promptstyle,caption={Prompt used to generate the Metric Estimation QA data (user message and rules/schema block). The JSON context is added at run time; the part-level version is the same, with object\_name replaced by part\_label.},label={lst:scale}]
You are generating FOUR Scale Estimator Q&A pairs for a robot-centric dataset.
Each Q&A must describe a size-related property of the given object.

Generate exactly FOUR Q&A pairs as follows:
1. One about the *overall size* (length x width x thickness)
2. One about *thickness*
3. One about *width*
4. One about *length*

Keep variety in phrasing across objects/frames, but the numeric values must be
copied exactly from the Compact Object Context. Do NOT invent or modify numbers.

The context includes an object_id for the labeled/bounded instance in the scene.
Use it ONLY internally to identify the correct instance; never mention an id or
label number. If multiple similar objects are visible, each question MUST specify
the target using natural spatial cues (left/right/middle, nearest/farthest,
"to the left of the table", "near the door"); never use numbers or coordinates.

In each ANSWER, wrap the object name with <p>...</p> immediately followed by <SEG>.
Do NOT include <p> or </p> in the questions.

--- Rules / schema block ---
Return ONLY valid JSON. Do not use Markdown or code fences. Keys in double quotes.
{
  "results": [
    { "question": "", "answer": "" },
    { "question": "", "answer": "" },
    { "question": "", "answer": "" },
    { "question": "", "answer": "" }
  ]
}
Rules:
- Use ONLY values in the context; units in meters with 2 decimals (e.g., 1.32 m).
- Output exactly four Q&A: overall size, thickness, width, length.
- Answers concise (1-2 sentences), physically consistent, robot-readable.
- The four questions must use four different starters.
Fields: object_name, object_id (internal only), size_m.length / width / thickness.

Example (structure only; do not copy wording):
{
  "results": [
    { "question": "Provide the overall size of the lamp stand near the wall.",
      "answer": "The <p>lamp stand</p><SEG> measures 1.25 m x 0.45 m x 0.40 m." },
    { "question": "Specify the thickness of the lamp stand closest to the camera.",
      "answer": "The thickness of the <p>lamp stand</p><SEG> is 1.25 m." }
  ]
}

Compact Object Context (copy numbers exactly; do not invent):
{ ... per-referent JSON appended here ... }
\end{lstlisting}

\begin{table}[H]
\centering
\caption{Per-task summary of the data-generation prompts. "per pair" means the prompt produces one QA item for every pair [i-j] of referents in the scene. All answers tag the grounded referent with \texttt{<p>...</p><SEG>}.}
\label{tab:deltas}
\footnotesize
\setlength{\tabcolsep}{4pt}
\resizebox{\linewidth}{!}{%
\begin{tabular}{@{}llcl@{}}
\toprule
Subtask & Context fields & \#QA & Notes \\
\midrule
Metric Estimation
  & \texttt{size\_m.\{l,w,thick\}}
  & 4
  & overall size, thickness, width, length \\
Grounded Measurement
  & \texttt{size\_m}, \texttt{dist\_fwd\_m}
  & 5
  & 4 size questions + 1 camera distance \\
Distance Queries
  & \texttt{relative\_distance[i-j]}
  & per pair
  & distance between the two referents \\
Spatial Relations
  & \texttt{relative\_position[i-j]}
  & per pair
  & left/right label; reverse flips direction \\
Depth Relations
  & \texttt{relative\_depth\_position[i-j]}
  & per pair
  & closer/farther; reverse flips label and order \\
Size Comparison
  & \texttt{size\_m}, \texttt{height},
  & per pair
  & winner taken from the ranking, \\
  & \texttt{desc\_order\_object\_size}
  &
  & \quad not from computed volume \\
Existence Verification
  & one real referent + a distractor
  & 2 / frame
  & only the "exists" answer carries \texttt{<SEG>} \\
Functional Grounding
  & precomputed instance context
  & varies
  & question states the intent, not the referent name \\
\bottomrule
\end{tabular}%
}
\end{table}